\documentclass[12pt, letterpaper]{article}

\usepackage[utf8]{inputenc}
\usepackage[T1]{fontenc}
\usepackage{amsmath}
\usepackage{amssymb}
\usepackage{booktabs}
\usepackage{float}
\usepackage{geometry}
\usepackage{setspace}
\usepackage{lineno} % for line number
\usepackage[style=apa, backend=biber]{biblatex}
\usepackage[colorlinks=true,allcolors=blue]{hyperref}
\usepackage{caption}
\usepackage{hyperref}
\usepackage{graphicx} %add [draft] before graphic to deactivate the figures

\begin{document}

%% Title block
\begin{center}
    {\LARGE\bfseries Boundary condition fidelity for bottom-hole pressure and CO\textsubscript{2} plume prediction in geological carbon storage}

    \vspace{18pt}

    {\normalsize
    Romal Ramadhan$^{1}$,\quad Seyyed A. Hosseini$^{1,2}$, \quad Larry W. Lake$^{3}$}

    \vspace{8pt}

    {\small $^{1}$Department of Earth and Planetary Sciences, Jackson School of Geosciences, The University of Texas at Austin, Austin, TX 78712, USA}\\[4pt]
    {\small $^{2}$Bureau of Economic Geology, The University of Texas at Austin, 10611 Exploration Way, Austin, TX 78758, USA}\\[4pt]
    {\small $^{3}$Hildebrand Department of Petroleum and Geosystems Engineering, The University of Texas at Austin, Austin, TX 78712, USA}\\[4pt]
    {\small $^{*}$Correspondence: romalramadhan@utexas.edu, seyyed.hosseini@beg.utexas.edu}

    \vspace{18pt}
\end{center}

\begin{abstract}
Accurate prediction of bottom-hole pressure (BHP) and CO$_2$ plume migration is essential for safe geological carbon storage, yet practical simulations often rely on truncated domains where artificial boundaries distort pressure diffusion and CO$_2$ saturation footprints. In this study, we evaluate how boundary-condition fidelity affects BHP and CO$_2$ plume prediction by comparing ten reduced-domain boundary treatments against full-domain reference simulations in homogeneous and heterogeneous reservoirs. We test uniform pore-volume multipliers, transmissibility modifiers, corner-adjusted pore-volume corrections, layered corrections, and gradual modifiers using BHP RMSE, NRMSE, peak pressure deviation, and plume Intersection over Union (IoU) as performance metrics. Our results show that conserving corner pore volume is the most important requirement for truncated-domain modeling. We find that uniform treatments which neglect corner storage generate large pressure errors, with BHP RMSE of 362 to 382 psi in the homogeneous model and 250 to 304 psi in the heterogeneous model, and yield plume IoU values near 0.80 to 0.84, indicating roughly 16 to 20\% of the combined plume area is misrepresented. Corner-adjusted scenarios substantially reduce pressure errors and raise plume IoU above 0.94, but we observe that transmissibility correction is not universally beneficial. In homogeneous reservoirs, uniform transmissibility adjustment improves pressure fidelity; in heterogeneous reservoirs, it can over-restrict flow across variable-permeability boundary faces, increasing BHP error and contracting the predicted plume. We find the gradual modifier with transmissibility correction provides the most consistent performance, achieving BHP NRMSE below 3.7\% and plume IoU above 0.97 in both reservoir types.

\vspace{6pt}

\end{abstract}

\vspace{12pt}
\hrule
\vspace{20pt}

%% ============================================================
\section{Introduction}
%% ============================================================
Accurate prediction of bottom-hole pressure (BHP) and carbon dioxide (CO\textsubscript{2}) plume migration is key to the safe and economic deployment of geological carbon storage (GCS) \parencite{TalabiRenMisra2026}. These two response variables span the coupled pressure and spatial dimensions of GCS risk: BHP governs injection rate design and the maintenance of injection pressures below formation fracture gradients, while plume extent defines the subsurface footprint within which displacement, dissolution, and trapping processes occur. Both carry direct regulatory weight \parencite{LuEtAl2022, WangHosseiniBump2025}. BHP serves as the primary indicator that operations remain within safe pressure limits and protect underground sources of drinking water (USDW), and plume extent delineates the area of review (AoR) under EPA Class VI permitting within which operators must identify and remediate potential leakage pathways \parencite{USEPA2013, SinghAtesVaidya2025}. During operations, deviations between simulated and observed BHP or plume behavior can trigger injection rate adjustments, additional monitoring obligations, or regulatory review, making simulation fidelity in both dimensions directly consequential to project economics and timeline \parencite{IEAGHG2020}.
 
Achieving that fidelity is complicated by the scale mismatch inherent in GCS simulation \parencite{MaoJahanbaniGhahfarokhi2024}. Pressure perturbations generated by CO\textsubscript{2} injection propagate tens to hundreds of kilometers from the well over decadal timescales, far exceeding the spatial extent that can be practically captured in a high-resolution numerical model, while the plume itself may migrate only hundreds of meters to several kilometers depending on permeability, heterogeneity, and buoyancy forces \parencite{BumpHovorka2024}. Domain truncation is therefore a practical necessity across nearly every stage of a GCS workflow, from early screening through detailed characterization and operational monitoring \parencite{RamadhanHosseini2026}. The cost of truncation, however, is the introduction of artificial boundaries that reflect pressure back toward the injection well rather than allowing it to dissipate into the far field. This reflection elevates simulated BHP above what an unbounded domain would produce and distorts plume trajectories by artificially constraining the pressure gradients that drive lateral migration.
 
Several modifier-based boundary treatments have been proposed to mitigate these artifacts. \textcite{GhomianEtAl2024} showed that pore-volume multiplier (PVM) and transmissibility modifier (TM) schemes applied at boundary cells can materially affect BHP and area-of-review forecasts, and that gradational modifier schemes improve agreement with full-domain reference behavior. \textcite{YangEtAl2025} similarly demonstrated that combining boundary-cell extension with equivalent PVM and TM offers further gains over simple fixed-pressure or excessively large pore-volume simplifications, particularly for regionally extensive open aquifers, with smaller but still relevant effects on plume migration. In our previous work, we introduced a gradual modifier framework for storage-conserving truncated models and demonstrated improved AoR prediction efficiency by distributing pore-volume and transmissibility corrections across boundary cells \parencite{RamadhanHosseini2026}. Despite this progress, existing studies have generally assessed pressure or plume behavior in isolation, and none has established whether a single boundary treatment preserves both simultaneously across reservoir types ranging from homogeneous to heterogeneous to full-scale models.
 
There is another source of error here that we have not previously pinned down. Corner cells, which sit where two domain edges meet, end up representing a surprisingly large chunk of reservoir volume that uniform modifier schemes simply miss. When we lose that corner volume, we also lose part of the pressure buffer the injector relies on, which pushes BHP higher and introduces gradient errors near the boundaries. Those errors do not stay put; they bleed into the plume migration paths the model predicts. One proposed fix is to grade the modifiers spatially, distributing the corrections as a power-law function of distance from the center of the domain. The idea is to recover the missing corner volume while smoothing out the sharp property contrasts that uniform schemes create. What we have not yet shown is whether this geometric correction actually delivers better predictions, both for BHP and for plume behavior, across the range of reservoir architectures we would expect to encounter in practice.
 
We address these gaps through an evaluation of boundary condition configurations against full-domain reference simulations in homogeneous and heterogeneous reservoir models. Using BHP time series and CO\textsubscript{2} plume extent as co-equal validation criteria, we quantify performance with peak pressure deviation, normalized root-mean-square error (NRMSE), and plume area metrics over the full injection and post-injection monitoring periods. The results provide practical guidance for boundary condition selection across GCS project stages, with direct implications for injection design, area-of-review delineation, and regulatory compliance demonstration.
 
%% ============================================================
\section{Methodology}
%% ============================================================
\subsection{Numerical model description}
\label{sec:model}
 
We construct reservoir models of increasing geological complexity to evaluate how boundary condition affects BHP and CO$_2$ plume prediction. A homogeneous model serves as a controlled baseline free of geological variability, while a heterogeneous model introduces spatial complexity. Both models share the same domain geometry and injection schedule so that we can attribute differences in BHP and plume behavior solely to the boundary treatment and reservoir heterogeneity. We perform all simulations in the CMG-GEM compositional simulator (Computer Modelling Group Ltd.) under isothermal conditions. We exclude geochemical reactions and geomechanical coupling to keep the dominant physics limited to pressure diffusion and buoyancy-driven CO$_2$ migration. Table~\ref{tab:properties} compiles the key model parameters.
 
\subsubsection{Homogeneous model}
\label{sec:homo}
 
We discretize the full-domain reference case, hereafter termed the \textit{truth} model, on a $501 \times 501 \times 1$ grid with cell dimensions of $113.65~\text{ft} \times 113.65~\text{ft} \times 328.1~\text{ft}$, covering a lateral area of $56{,}939~\text{ft} \times 56{,}939~\text{ft}$. The formation top lies at a depth of 3{,}281~ft with a uniform thickness of 328.1~ft. We initialize the reservoir pressure at 1{,}600~psi, and hold porosity and permeability constant at 0.20 and 10~mD throughout the grid. A centrally located injector delivers CO$_2$ at a surface-equivalent rate of 2{,}500~tonnes/day for 10~years, after which we shut in the well and monitor the system for a further 40~years.
 
We extract two truncated variants of $251 \times 251 \times 1$ and $101 \times 101 \times 1$ cells concentrically around the injection well. All interior cells retain the same properties as the truth; we modify only the outermost cell ring according to the boundary treatments.
 
\subsubsection{Heterogeneous model}
\label{sec:hetero}
 
\begin{figure}[!ht]
\centering
\includegraphics[width=\textwidth]{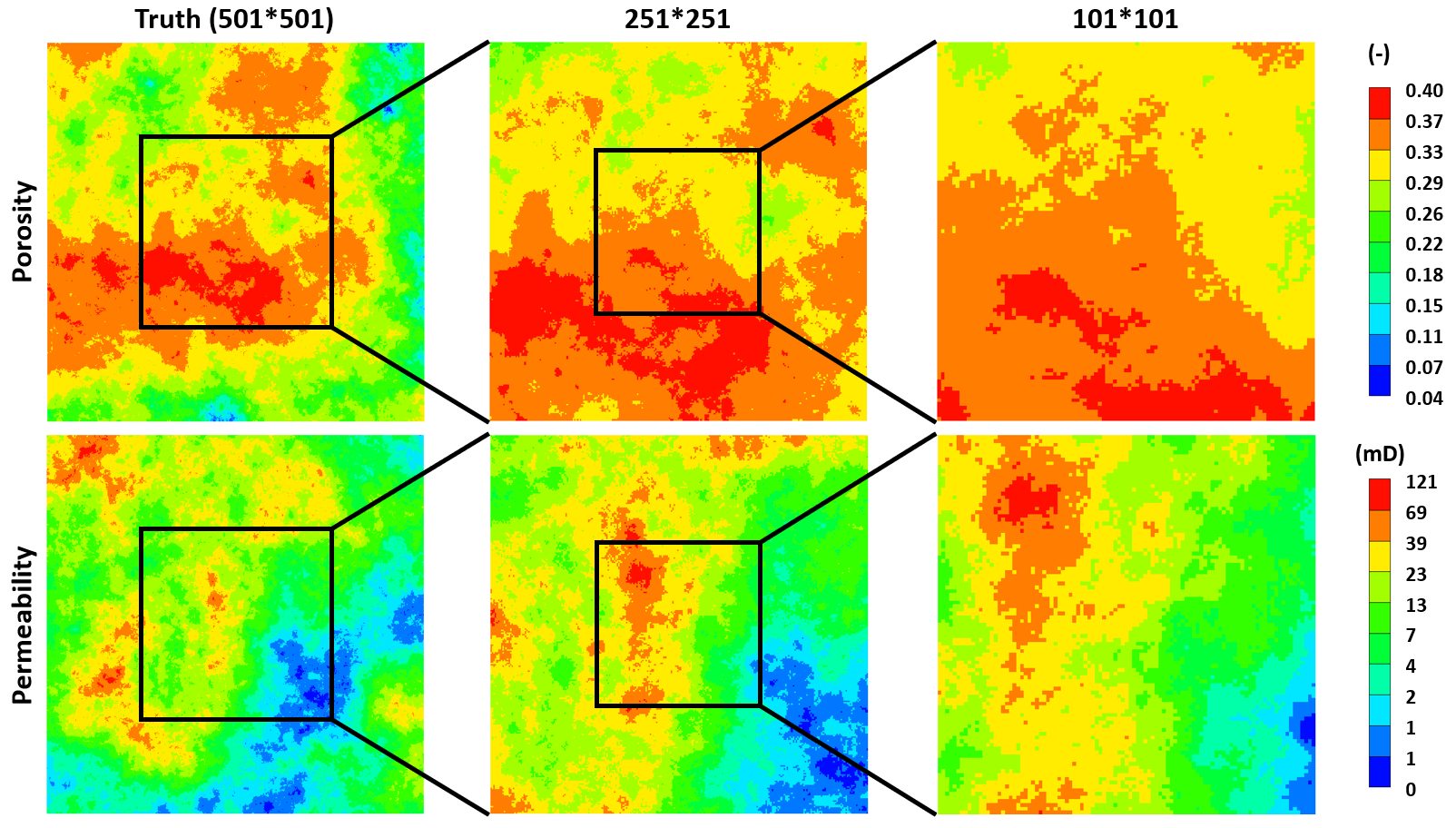}
\caption{Spatial distributions of porosity (top row) and permeability (bottom row, logarithmic scale) for the heterogeneous reservoir. The full-domain truth model ($501 \times 501$) is shown on the left, with the $251 \times 251$ and $101 \times 101$ reduced-domain extractions in the center and right panels. Black outlines and connecting lines indicate the nested sub-domain boundaries (modified from \cite{RamadhanHosseini2026}).}
\label{fig:Figure1}
\end{figure}

To generate spatially correlated porosity and permeability fields on the full $501 \times 501 \times 1$ grid, we implement sequential Gaussian simulation. We then rescale the realizations so that their arithmetic averages coincide with the homogeneous values (0.20 porosity, 10~mD permeability). Figure~\ref{fig:Figure1} displays the resulting porosity and permeability distributions for the truth model and the two reduced-domain extractions. Reduced-domain sub-models of $251 \times 251 \times 1$ and $101 \times 101 \times 1$ cells are carved from the truth grid using the \texttt{NULL} keyword, which deactivates cells outside the region of interest while preserving the full-resolution property fields inside the active domain. We apply the same boundary-layer modifications as in the homogeneous cases; all other simulation settings remain identical to the truth.
 
\begin{table}[!ht]
\centering
\caption{Reservoir model properties for the full domain simulation (after \cite{RamadhanHosseini2026}).}
\label{tab:properties}
\begin{tabular}{l c c}
\toprule
\textbf{Parameter} & \textbf{Value} & \textbf{Unit} \\
\midrule
Grid dimensions                 & $501 \times 501 \times 1$                      & --              \\
Cell dimensions                 & $113.65 \times 113.65 \times 328.1$            & ft              \\
Domain area                     & $56{,}939 \times 56{,}939$                     & ft$^2$          \\
Top depth                       & 3{,}281                                        & ft              \\
Formation thickness             & 328.1                                          & ft              \\
Initial reservoir pressure      & 1{,}600                                        & psi             \\
Porosity (homo / hetero mean)   & 0.20                                           & fraction        \\
Permeability (homo / hetero mean) & 10                                           & mD              \\
CO$_2$ injection rate           & 2{,}500                                        & tonnes/day      \\
Injection period                & 10                                             & years           \\
Post-injection monitoring       & 40                                             & years           \\
\bottomrule
\end{tabular}
\end{table}

\subsection{Boundary condition formulation}
\label{sec:bc}
 
We scale each boundary cell's pore volume by a multiplier that accounts for the reservoir volume removed beyond the grid edge. For the $251 \times 251$ and $101 \times 101$ grids, we set the uniform PVM values to 126 and 201 (later on, we call this as $c$), respectively, so that the aggregate boundary-cell pore volume equals that of the excised portion of the full domain \cite{RamadhanHosseini2026}. We then compute transmissibility between two adjacent cells from the harmonic mean of their individual half-cell transmissibilities. For face-center transmissibilities in the $j$-direction on the outer boundary \parencite{GhomianEtAl2024}:

\begin{align}
T_{j+1/2} &= \frac{K_a \left( \Delta x_a \cdot \Delta z_a \right)}{\Delta y_a / 2} \label{eq:half_trans_a} \\[6pt]
T_{j-1/2} &= \frac{K_b \left( \Delta x_b \cdot \Delta z_b \right)}{\Delta y_b / 2} \label{eq:half_trans_b}
\end{align}
where $K_a$ and $K_b$ are the cell permeabilities, and $\Delta x$, $\Delta y$, $\Delta z$ are the cell dimensions in the $x$, $y$, and $z$ directions. The harmonic-mean transmissibility at the interface is:
\begin{equation}
T_j = \frac{1}{\dfrac{1}{T_{j+1/2}} + \dfrac{1}{T_{j-1/2}}}
\label{eq:harm_trans}
\end{equation}
 
When we impose a PVM on a boundary cell, the transmissibility must be adjusted to reflect the modified cell geometry. In the $i$ and $k$ directions, where both cells sharing the interface carry the same volume scaling, the adjusted half-cell transmissibilities become:
\begin{align}
T_{i+1/2}^{\,\mathrm{adj}} &= \frac{K_b \left( \mathrm{PVM} \cdot \Delta y \cdot \Delta z \right)}{\Delta x / 2} \label{eq:adj_ik_a} \\[6pt]
T_{i-1/2}^{\,\mathrm{adj}} &= \frac{K_c \left( \mathrm{PVM} \cdot \Delta y \cdot \Delta z \right)}{\Delta x / 2} \label{eq:adj_ik_b}
\end{align}
Because PVM factors through both terms symmetrically, the transmissibility modifier in the $i$ and $k$ directions reduces to:
\begin{equation}
\mathrm{TM}_{i,k} = \frac{T_i^{\,\mathrm{adj}}}{T_i} = \mathrm{PVM}
\label{eq:tm_ik}
\end{equation}
 
In the $j$-direction, the interface connects a modified boundary cell (PVM applied) to an unmodified interior cell ($\mathrm{PVM} = 1$), producing an asymmetric configuration. The adjusted half-cell transmissibility on the boundary side is:
\begin{equation}
T_{j-1/2}^{\,\mathrm{adj}} = \frac{K_b \left( \Delta x_b \cdot \Delta z_b \right)}{\mathrm{PVM} \cdot \Delta y_b / 2} = \frac{T_{j-1/2}}{\mathrm{PVM}}
\label{eq:adj_j}
\end{equation}
Harmonic averaging across the dissimilar cells then yields a transmissibility modifier of:
\begin{equation}
\mathrm{TM}_j = \frac{T_j^{\,\mathrm{adj}}}{T_j} = \frac{2}{\mathrm{PVM} + 1}
\label{eq:tm_j}
\end{equation}

\begin{figure}[!ht]
\centering
\includegraphics[width=0.85\textwidth]{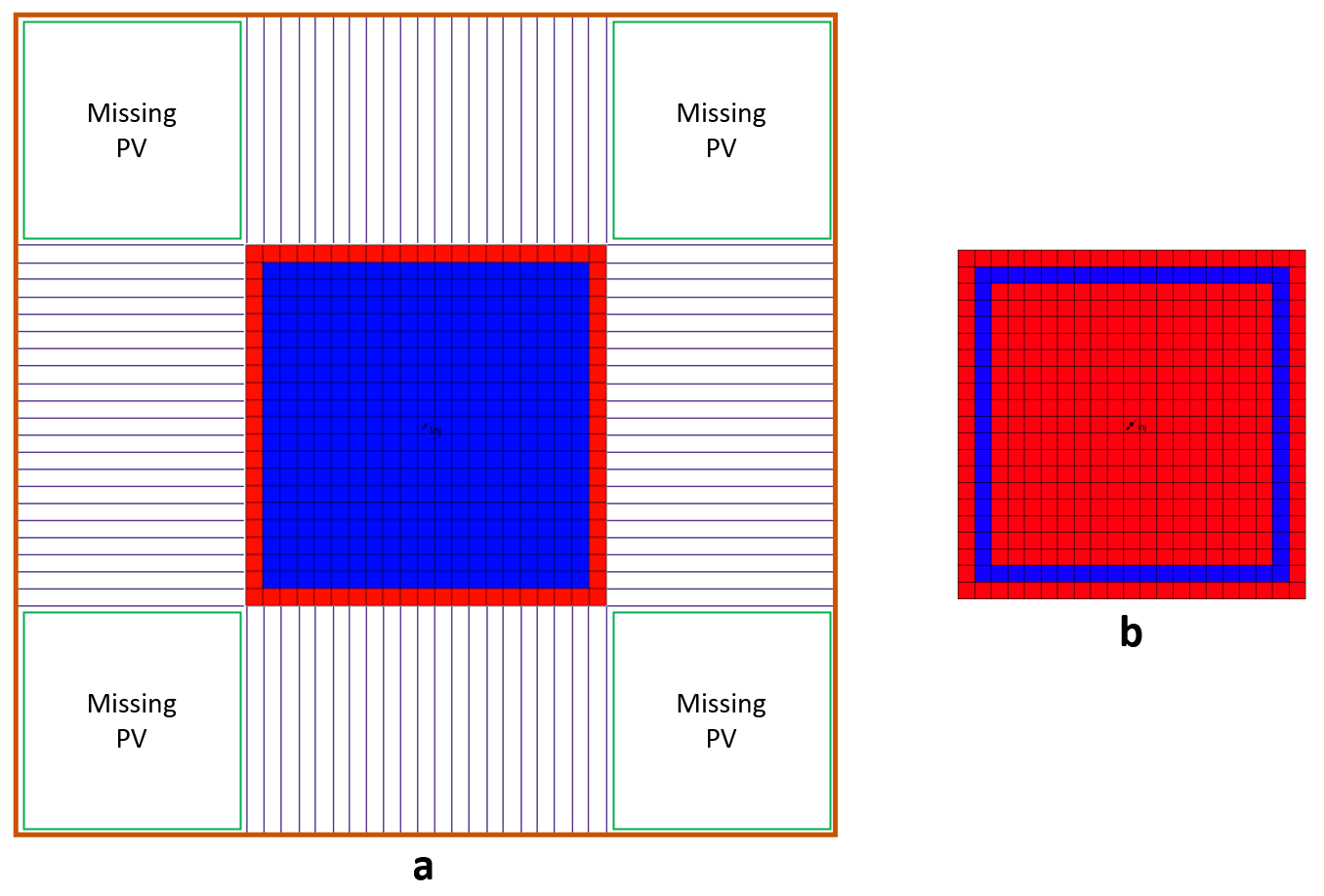}
\caption{Illustration of the corner pore-volume deficit in conventional boundary-cell modification: (a) the uniform PVM scheme applied to the outermost cell ring (red) leaves the four corner regions unaccounted for (labeled ``Missing PV'')(b) the corresponding transmissibility adjustment on the outer cell faces (modified after \cite{RamadhanHosseini2026}).}
\label{fig:Figure2}
\end{figure}

Cells located at the corners of the domain are affected by two truncated edges at once. Because of this, they represent a much larger share of the missing reservoir volume than cells along a single boundary, and a uniform PVM correction does not fully recover that loss. Figure~\ref{fig:Figure2}a shows that edge-only scaling still leaves the four corner regions undercompensated, while Figure~\ref{fig:Figure2}b identifies the transmissibility faces that must be adjusted. To correct this remaining deficit, the corner-adjusted schemes (VA and VAT) apply an additional factor of $c^2$ to the PVM of each corner cell, where $c$ is the base multiplier. This adjustment ensures that the reduced model preserves the total pore volume of the truth model.

\subsection{Boundary condition configurations}
\label{sec:cases}
 
We test ten boundary condition scenarios on the reduced grids of $101 \times 101$ and $251 \times 251$ cells, alongside the $501 \times 501$ truth model, for both the homogeneous and heterogeneous reservoirs. Table~\ref{tab:scenarios} catalogs all cases. The simplest treatment applies a uniform PVM to the outermost cell layer only (\textbf{V}), or pairs it with the corresponding TM corrections (\textbf{VT}). The corner-adjusted variants (\textbf{VA}, \textbf{VAT}) are identical to V and VT but amplify the corner-cell pore volumes to recover the missing corner storage. The layered approach of \parencite{GhomianEtAl2024} distributes PVM and TM across four concentric cell layers from the domain edge inward (\textbf{GhV}, \textbf{GhVT}). In our implementation, we additionally adjust corner volumes to conserve total formation storage.
 
\begin{figure}[!ht]
\centering
\includegraphics[width=0.85\textwidth]{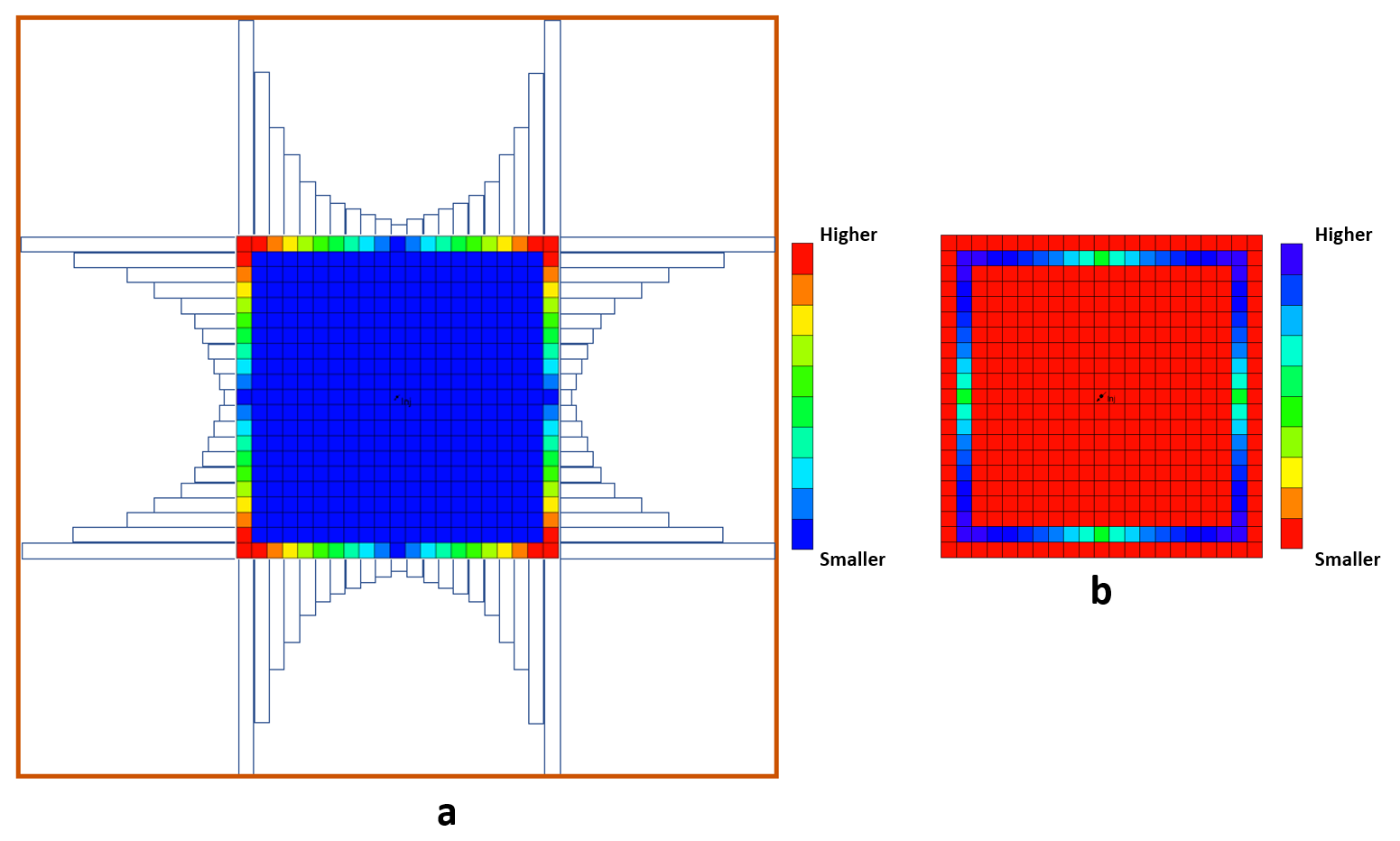}
\caption{Gradual boundary-cell modification scheme: (a) the PVM distribution increases progressively from the midpoint of each boundary edge toward the corners, shown as bar heights representing modifier magnitude, with the orange outline denoting the full-domain extent; (b) the corresponding plan-view representation of the gradual TM applied to the face-center outermost cell, (modified from \cite{RamadhanHosseini2026}).}
\label{fig:Figure3}
\end{figure}

\begin{figure}[!ht]
\centering
\includegraphics[width=0.85\textwidth]{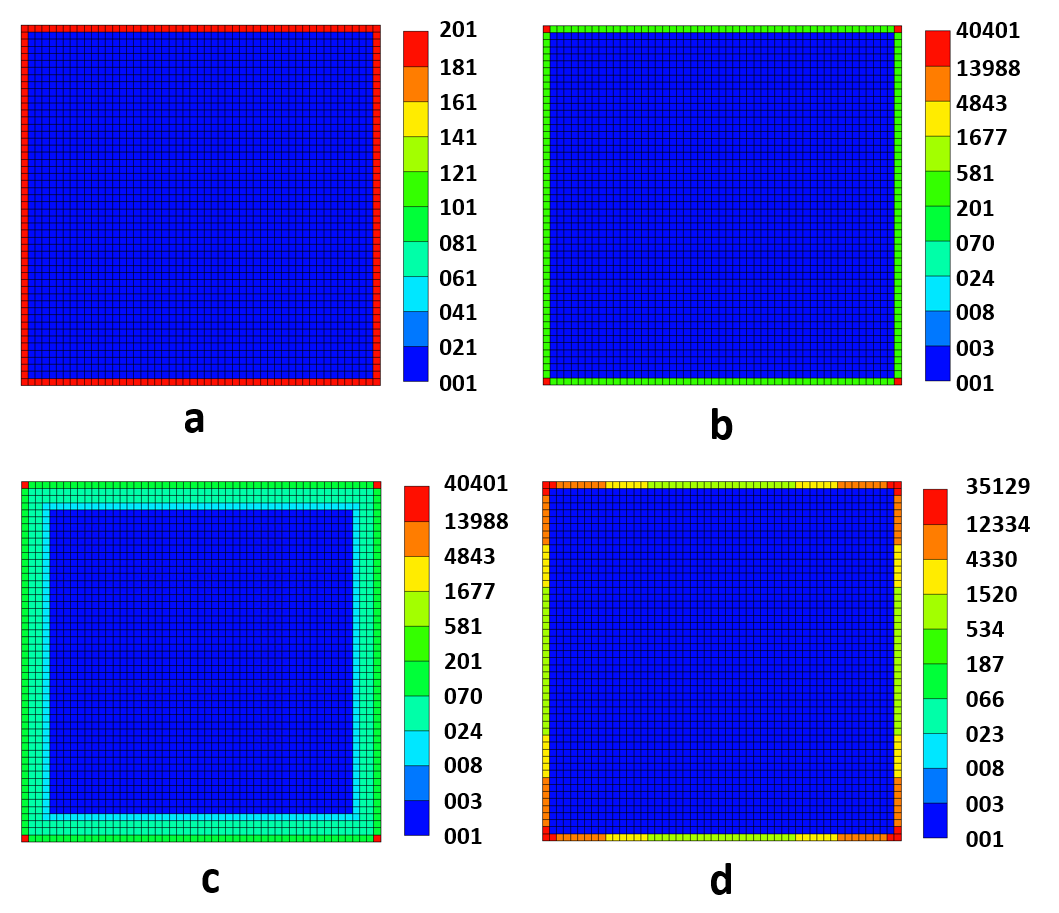}
\caption{Plan-view visualization of pore-volume modifier distributions for the four boundary treatment categories on the $101 \times 101$ grid: (a) uniform PVM applied to the outermost layer (cases V and VT), (b) uniform PVM with corner-cell volume correction (cases VA and VAT), (c) layered PVM following \textcite{GhomianEtAl2024} with corner adjustment (cases GhV and GhVT), and (d) gradual power-law PVM with progressive mid-side to corner transition (cases VAG and VAGT). Color scales indicate the modifier magnitude assigned to each boundary cell (after \cite{RamadhanHosseini2026}).}
\label{fig:Figure4}
\end{figure}

Our gradual power-law modifier \parencite{RamadhanHosseini2026} varies modifier magnitudes continuously along the single outermost cell ring (\textbf{VAG}, \textbf{VAGT}). The distributions follow:
\begin{align}
\mathrm{PVM}_i &= \mathrm{PVM}_{\min} + k \left(\frac{d_i}{d_{\max}}\right)^{b} \label{eq:pvm_grad} \\[6pt]
\mathrm{TM}_i &= \mathrm{TM}_{\min} + k \left(\frac{d_i}{d_{\max}}\right)^{b} \label{eq:tm_grad}
\end{align}
Here $d_i$ is the distance of boundary cell $i$ from the domain center, $d_{\max}$ is half the domain width, $\mathrm{PVM}_{\min}$ and $\mathrm{TM}_{\min}$ are the minimum modifier values at the midpoint of each edge, $b$ is the power-law exponent, and $k$ is a scaling coefficient that we determine by requiring the cumulative boundary-cell pore volume to match the volume removed by truncation. The large exponent keeps modifiers nearly flat across the central portion of each edge and concentrates the sharpest increase near the corners, recovering the missing corner storage without introducing abrupt property jumps. Figure~\ref{fig:Figure3}a depicts the resulting PVM distribution as bar heights that grow from the edge midpoints toward the corners, while Figure~\ref{fig:Figure3}b shows the corresponding plan-view representation on TM implementation. VAG applies only the pore-volume gradient; VAGT adds the transmissibility gradient. Figure~\ref{fig:Figure4} provides a side-by-side comparison of the PVM distributions for all four boundary treatment categories on the $101 \times 101$ grid.

\begin{table}[!ht]
\centering
\caption{Simulation scenario definitions and grid dimensions (after \cite{RamadhanHosseini2026}).}
\label{tab:scenarios}
\begin{tabular}{c l l c c c}
\toprule
\textbf{Case} & \textbf{Label} & \textbf{Description} & $\mathbf{501 \times 501}$ & $\mathbf{101 \times 101}$ & $\mathbf{251 \times 251}$ \\
\midrule
01 & Truth    & Full-domain reference              & $\checkmark$ &              &              \\
02 & V101     & Uniform PVM                        &              & $\checkmark$ &              \\
03 & VT101    & Uniform PVM + TM                   &              & $\checkmark$ &              \\
04 & VA101    & PVM, corner adjusted               &              & $\checkmark$ &              \\
05 & VAT101   & PVM + TM, corner adjusted          &              & $\checkmark$ &              \\
06 & GhV101   & \textcite{GhomianEtAl2024}\ PVM    &              & $\checkmark$ &              \\
07 & GhVT101  & \textcite{GhomianEtAl2024}\ PVM + TM &              & $\checkmark$ &              \\
08 & VAG101   & Gradual PVM                        &              & $\checkmark$ &              \\
09 & VAGT101  & Gradual PVM + TM                   &              & $\checkmark$ &              \\
10 & VAG251   & Gradual PVM                        &              &              & $\checkmark$ \\
11 & VAGT251  & Gradual PVM + TM                   &              &              & $\checkmark$ \\
\bottomrule
\end{tabular}
\end{table}
 
\subsection{Performance metrics}
\label{sec:metrics}
 
Each boundary configuration is assessed along two dimensions: pressure response and CO$_2$ saturation behavior. We compute the metrics defined below at every output time step, covering both the injection period and the post-injection monitoring window.
 
\subsubsection{Bottom-hole pressure}
\label{sec:bhp_metrics}
 
The peak pressure deviation is defined as the maximum absolute difference between the BHP of the reduced-domain model and the truth model. We report this deviation in psi and also express it as a percentage of the truth-model peak pressure:
\begin{equation}
\Delta \mathrm{BHP}_{\mathrm{peak}} = \max_{t}\,\bigl|\,\mathrm{BHP}_{\mathrm{reduced}}(t) - \mathrm{BHP}_{\mathrm{truth}}(t)\,\bigr|
\label{eq:peak_bhp}
\end{equation}
This metric captures the worst-case pressure error, which governs whether simulated injection pressures remain below formation fracture gradients.
 
We quantify the time-averaged BHP mismatch using the root-mean-square error (RMSE):

\begin{equation}
    \text{RMSE} = \sqrt{\frac{1}{N}\sum_{i=1}^{N}\left[\text{BHP}_{\text{reduced}}(t_i) - \text{BHP}_{\text{truth}}(t_i)\right]^2}
\end{equation}

\noindent and its normalized form (NRMSE), scaled by the dynamic range of the truth series:

\begin{equation}
    \text{NRMSE} = \frac{1}{\text{BHP}_{\max} - \text{BHP}_{\min}}\sqrt{\frac{1}{N}\sum_{i=1}^{N}\left[\text{BHP}_{\text{reduced}}(t_i) - \text{BHP}_{\text{truth}}(t_i)\right]^2}
\end{equation}

\noindent where $N$ is the number of output time steps and $\text{BHP}_{\max}$, $\text{BHP}_{\min}$ are the extremes of the truth BHP time series. NRMSE provides a dimensionless measure of how closely the reduced model reproduces the entire pressure trajectory, from transient build-up during injection through decline after shut-in.

\subsubsection{CO$_2$ plume extent}
\label{sec:metric_plume}

Spatial match between the reduced-domain and truth-model CO$_2$ plumes is quantified using the Intersection over Union (IoU), also known as the Jaccard index. Unlike a scalar area comparison, the IoU evaluates whether the reduced model occupies the same physical region as the truth, penalizing both spatial displacement and area mismatch simultaneously. Letting $\Omega_{\text{reduced}}(t)$ and $\Omega_{\text{truth}}(t)$ denote the sets of grid cells satisfying the gas-saturation threshold $S_g > 0.001$ at time $t$, we define the IoU as:

\begin{equation}
\text{IoU}(t) = \frac{|\Omega_{\text{reduced}}(t) \cap \Omega_{\text{truth}}(t)|}{|\Omega_{\text{reduced}}(t) \cup \Omega_{\text{truth}}(t)|}
\label{eq:iou}
\end{equation}

where $|\cdot|$ denotes the cumulative area (or cell count weighted by cell area) of the indicated set. The numerator measures the area shared by both plumes, and the denominator measures the area covered by either plume. By construction, $\text{IoU} \in [0, 1]$: a value of unity indicates perfect spatial coincidence, while a value of zero indicates complete spatial separation.

The IoU captures three distinct error modes in a single metric. Area underprediction reduces the intersection while leaving the union largely unchanged, lowering the IoU. Area overprediction inflates the union while the intersection remains capped at the truth area, again lowering the IoU. Lateral displacement of the plume, such as the redirection induced by boundary-pressure artifacts interacting with heterogeneous permeability fields, reduces the intersection and inflates the union simultaneously, producing a steeper penalty than either area-only error mode. We choose this metric because its joint sensitivity to area and location makes it well suited to heterogeneous reservoirs, where the plume front follows preferential flow paths and the relevant fidelity question is whether the reduced model places CO$_2$ in the correct location rather than merely whether it reproduces the correct total footprint.

%% ============================================================
\section{Results and Discussion}
%% =========================================================
\subsection{Bottom-hole Pressure Prediction}

\subsubsection{Homogeneous}

\begin{figure}[htbp]
    \centering
    \includegraphics[width=\textwidth]{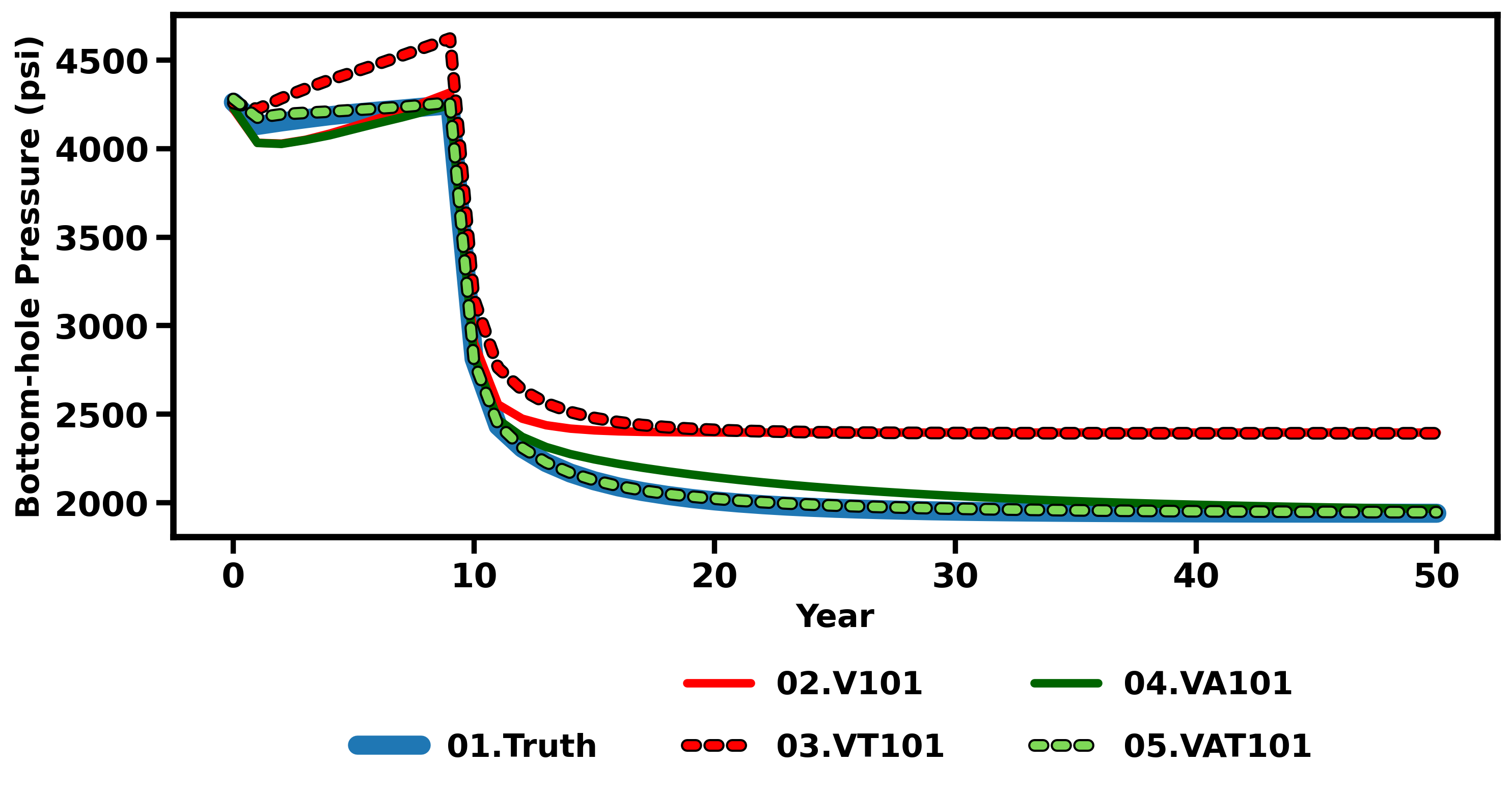}
    \caption{BHP comparison for the uniform and corner-adjusted boundary treatments on the
    $101 \times 101$ grid (homogeneous model).}
    \label{fig:Figure5}
\end{figure}

The uniform and corner-adjusted boundary treatments are first evaluated on the $101 \times 101$ grid (Figure~\ref{fig:Figure5}). All models capture the expected pressure response: a gradual injection-driven build-up from approximately 4,100 to 4,250~psi over 10~years, followed by post-injection pressure decline toward an equilibrium value near 1,940~psi. However, the size of the error differs substantially across boundary configurations, with deviations spanning more than an order of magnitude.

The two uniform PVM cases produce the largest  relative to the truth. V101 underpredicts during early injection (up to $-127$ psi at year 2), crosses the truth near year 8, and then overpredicts persistently after shut-in, stabilizing $+455$ psi above the truth at year 50. VT101 overpredicts throughout injection because the transmissibility modifier restricts the drainage, elevating BHP progressively from $+61$ psi at year 1 to $+393$ psi at year 10. Despite their contrasting injection-phase behavior, V101 and VT101 converge to nearly identical long-term plateaus ($\sim$2,395 psi), confirming that the persistent post-injection error is governed by the total boundary pore volume rather than by transmissibility: both cases lack the corner-volume correction needed to match the true far-field storage. Their RMSE values are 362.14 psi (NRMSE = 11.60\%) for V101 and 382.35 psi (NRMSE = 12.25\%) for VT101 (Table~\ref{tab:rmse_homo}). The slightly larger error in VT101 indicates that adding a transmissibility modifier on top of an already deficient pore-volume distribution amplifies, rather than alleviates, the pressure inflation during injection.

Corner adjustment substantially improves performance. VA101 retains the early injection underprediction common to all PVM-only schemes ($-128$ psi at year 2) but reduces the terminal offset to 25 psi at year 50, cutting the RMSE to 139.26 psi (NRMSE = 4.46\%), a reduction of approximately 62\% relative to V101. Pairing the corner correction with the transmissibility modifier (VAT101) further compresses both the early-injection underprediction and the late-time drift, lowering the RMSE to 108.70 psi (NRMSE = 3.48\%). The incremental benefit of the transmissibility modifier over corner-adjusted PVM alone is modest in absolute terms, about 30 psi in RMSE, but consistent in direction: once pore volume is correctly distributed, restricting transmissibility into the enlarged boundary cells brings the simulated pressure trajectory closer to the truth.

\begin{figure}[htbp]
    \centering
    \includegraphics[width=\textwidth]{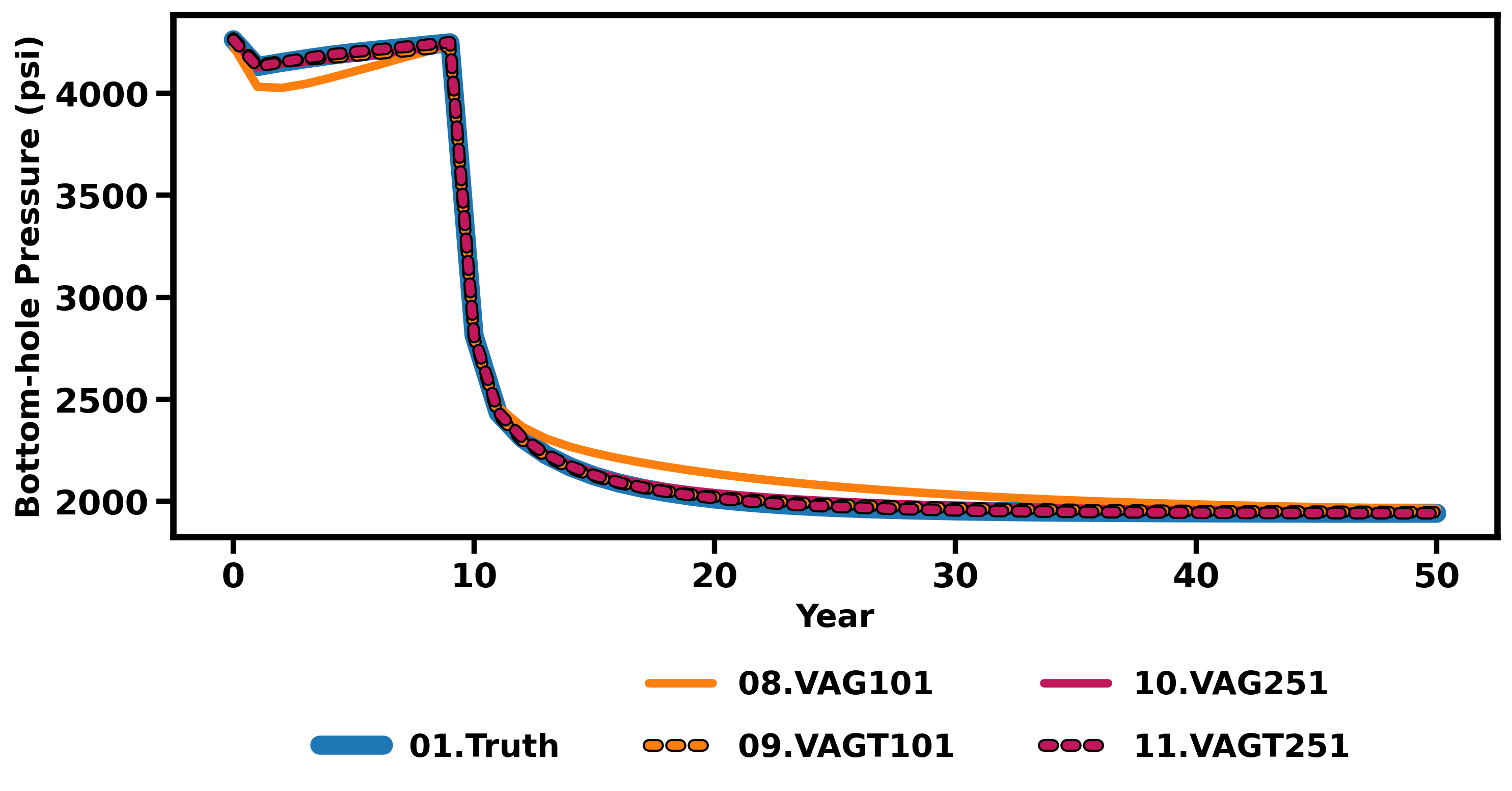}
    \caption{BHP comparison for the gradual power-law boundary treatments at two different number of grid used in simulation (homogeneous model).}
    \label{fig:Figure6}
\end{figure}

The effect of number of grid cells on the gradual modifier is evident in the comparison of the $101 \times 101$ and $251 \times 251$ implementations (Figure~\ref{fig:Figure6}). On the lesser grid, VAG101 exhibits the characteristic PVM-only underprediction during injection ($-128$ psi at year 2) and post-injection overshoot ($+126$ psi at year 18), with an RMSE of 137.13 psi (NRMSE = 4.39\%). Adding the transmissibility gradient (VAGT101) reduces the RMSE to 111.07 psi (NRMSE = 3.56\%). Enlarging the domain to $251 \times 251$ cells produces only a marginal further improvement: VAG251 yields an RMSE of 111.41 psi (NRMSE = 3.57\%) and VAGT251 reaches 110.11 psi (NRMSE = 3.53\%). The near-equivalence between the 101-cell VAGT and the 251-cell suggests that, in a homogeneous reservoir, the gradual modifier extracts most of the available accuracy already at the lesser grid; enlarged area cells do not materially expand the pressure buffer between the injection well and the boundary because the underlying pressure-diffusion is already well represented.

\begin{figure}[htbp]
    \centering
    \includegraphics[width=\textwidth]{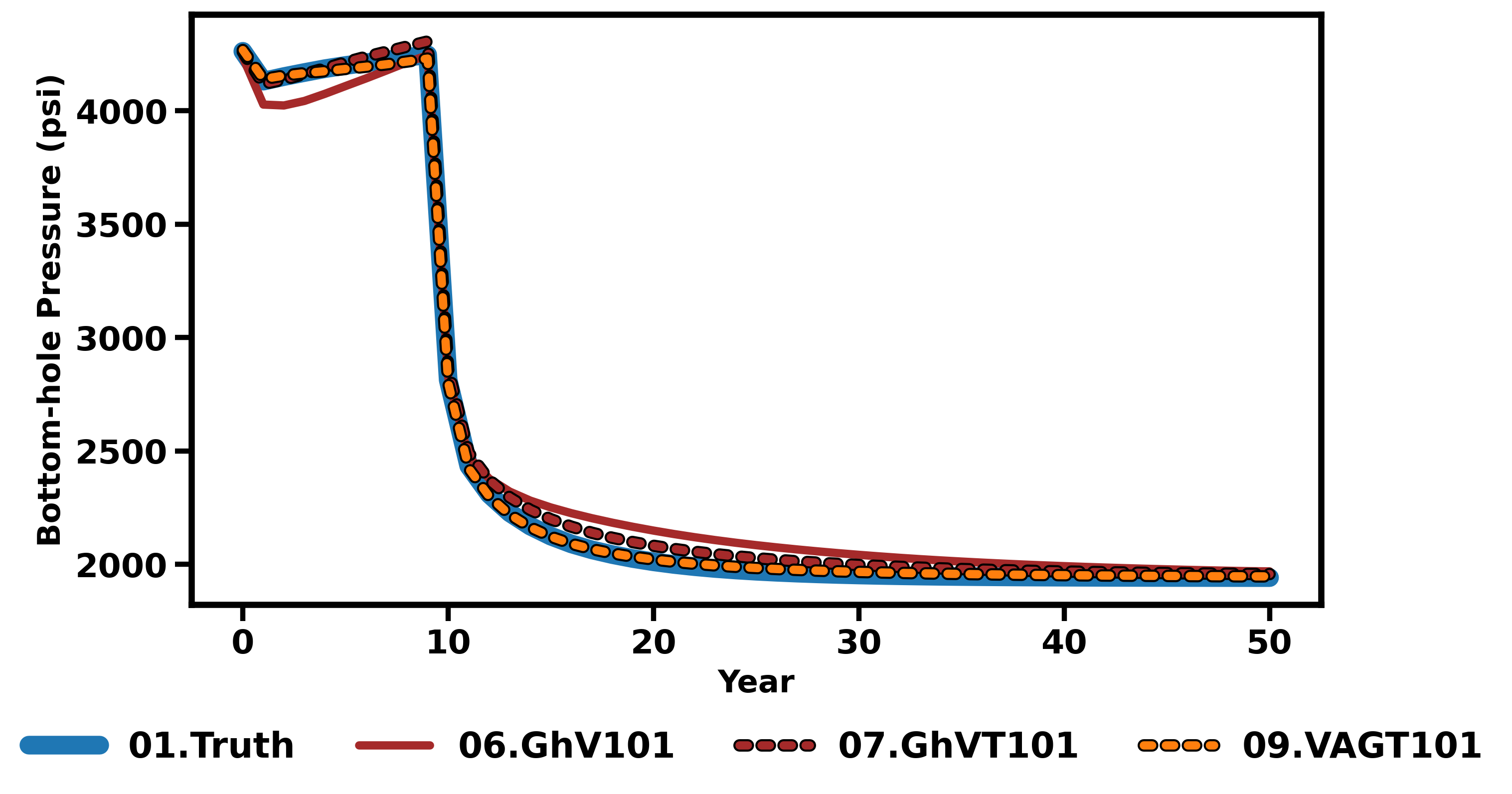}
    \caption{BHP comparison for the layered (GhV101, GhVT101) and gradual (VAGT101)
    boundary treatments on the $101 \times 101$ grid (homogeneous model).}
    \label{fig:Figure7}
\end{figure}

The layered scheme of \parencite{GhomianEtAl2024} is next compared with the gradual power-law modifier on the $101 \times 101$ grid (Figure~\ref{fig:Figure7}). GhV101, which distributes PVM across four concentric layers with corner adjustment, follows an injection-phase trajectory similar to VA101 (up to $-133$ psi underprediction) and overshoots by 141 psi near year 18 before converging, yielding an RMSE of 140.98 psi (NRMSE = 4.52\%). Adding the transmissibility correction (GhVT101) reduces the RMSE to 113.97 psi (NRMSE = 3.65\%), although a post-injection overshoot of approximately 76 psi persists near year 16.

The VAGT101 trajectory tracks the truth within a comparable envelope, with the peak deviation occurring during mid-injection and a terminal offset at year 50 of only a few psi. Its RMSE of 111.07 psi (NRMSE = 3.56\%) is slightly lower than that of GhVT101, consistent with the continuous power-law gradation distributing the boundary correction more smoothly than the four-layer step function. The three corner-adjusted schemes with transmissibility correction, VAT101, GhVT101, and VAGT101, therefore cluster within a narrow band of 108.70 to 113.97 psi RMSE on the $101 \times 101$ grid, indicating that once corner-volume conservation is achieved, the specific spatial distribution of the modifier exerts only a secondary influence on BHP fidelity in the homogeneous case.

All configurations are summarized in Table~\ref{tab:rmse_homo}. Two findings are consistent across treatments. The dominant correction is corner-volume conservation, which reduces RMSE by roughly 60\% (V101 to VA101) to 72\% (VT101 to VAT101) relative to the corresponding uniform variant. A secondary effect, transmissibility adjustment, provides a further reduction of approximately 20 to 30~psi in RMSE, but only once the total pore volume is correctly distributed; applied on top of a deficient uniform PVM, as in VT101, it worsens rather than improves the match. Among the $101 \times 101$ configurations, VAT101 yields the lowest RMSE (108.70~psi), with VAGT101 (111.07~psi) and GhVT101 (113.97~psi) falling within about 5~psi of that value. The 251-cell gradual variants offer no meaningful additional accuracy gain in the homogeneous reservoir, an observation that sets up a contrasting expectation for the heterogeneous case, where the distributed-correction strategies are likely to prove more discriminating.

\begin{table}[htbp]
    \centering
    \caption{RMSE and NRMSE of BHP predictions for all boundary condition configurations
    on the homogeneous model.}
    \label{tab:rmse_homo}
    \begin{tabular}{lcc}
        \hline
        \textbf{Model} & \textbf{RMSE (psi)} & \textbf{NRMSE (\%)} \\
        \hline
        V101     & 362.14 & 11.60 \\
        VT101    & 382.35 & 12.25 \\
        VA101    & 139.26 &  4.46 \\
        VAT101   & 108.70 &  3.48 \\
        GhV101   & 140.98 &  4.52 \\
        GhVT101  & 113.97 &  3.65 \\
        VAG101   & 137.13 &  4.39 \\
        VAGT101  & 111.07 &  3.56 \\
        VAG251   & 111.41 &  3.57 \\
        VAGT251  & 110.11 &  3.53 \\
        \hline
    \end{tabular}
\end{table}

\subsubsection{Heterogeneous}

We repeat the BHP comparison on the heterogeneous reservoir, where spatially correlated porosity and permeability fields introduce local flow-path variability that the boundary treatment must accommodate in addition to the far-field storage deficit. The truth model builds from approximately 2,850 psi to 3,083 psi over the 10-year injection period and declines to roughly 1,921 psi by year 50, a dynamic range that is smaller than in the homogeneous case owing to the lower effective transmissibility of the heterogeneous field.

\begin{figure}[htbp]
    \centering
    \includegraphics[width=\textwidth]{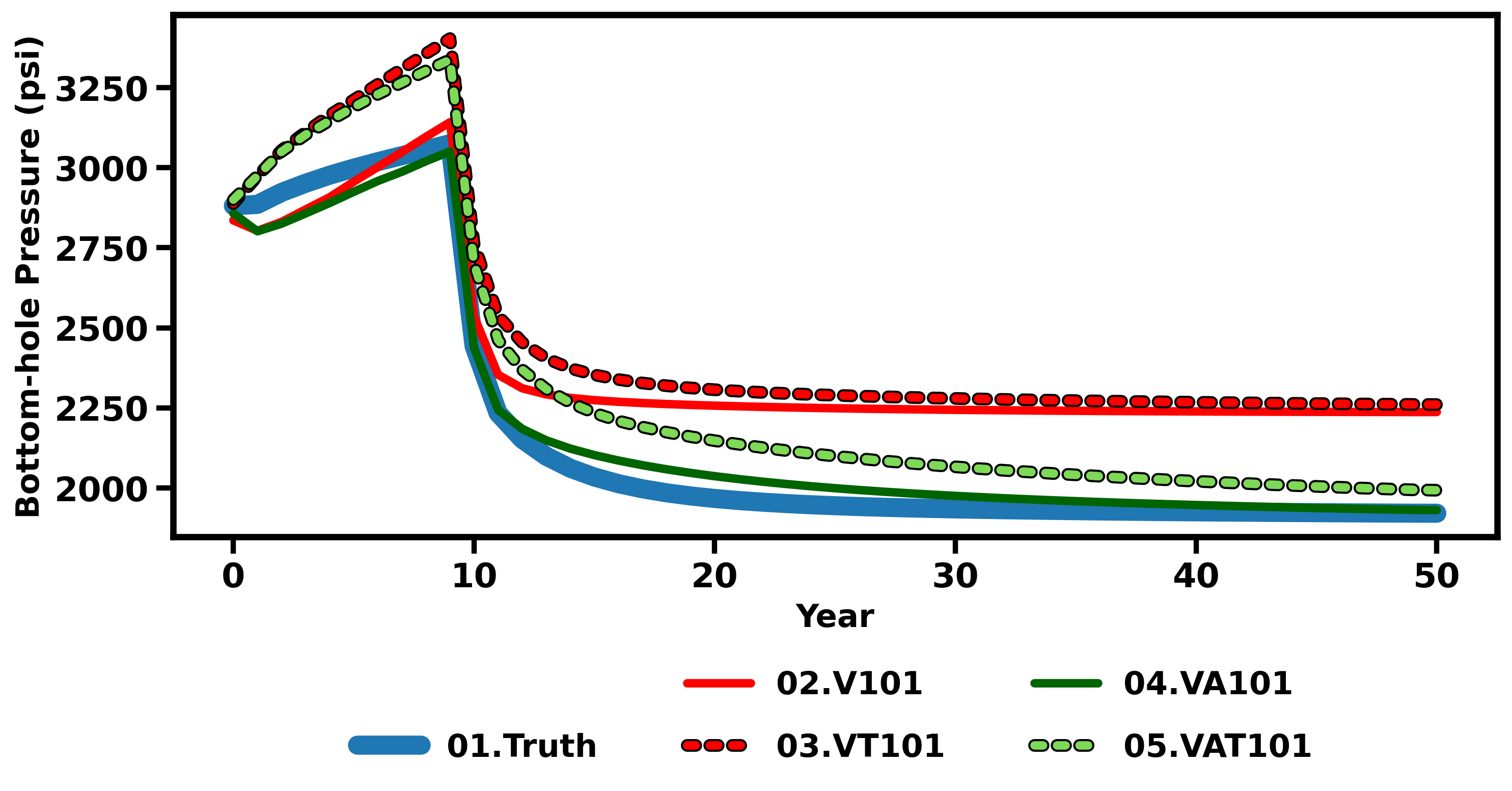}
    \caption{BHP comparison for the uniform and corner-adjusted boundary treatments on the
    $101 \times 101$ grid (heterogeneous model).}
    \label{fig:Figure8}
\end{figure}

The uniform PVM cases again produce the largest deviations (Figure~\ref{fig:Figure8}). V101 underpredicts during early injection (up to $-92$ psi at year 1), crosses the truth near year 7, and overpredicts persistently after shut-in, stabilizing 316 psi above the truth at year 50. VT101 overpredicts throughout injection, with the deviation growing from $+117$ psi at year 1 to $+343$ psi at year 9, and settles at a similar long-term plateau roughly 340 psi above the truth. These patterns mirror those in the homogeneous model, confirming that the failure of the uniform PVM to recover corner-cell storage is geometry-driven and largely independent of reservoir heterogeneity. Their RMSE values are 250.11 psi (NRMSE = 17.63\%) for V101 and 304.17 psi (NRMSE = 21.44\%) for VT101 (Table~\ref{tab:rmse_hetero}). As in the homogeneous case, the larger error in VT101 indicates that layering a transmissibility modifier onto a deficient pore-volume distribution magnifies, rather than mitigates, the pressure inflation.

A notable shift from the homogeneous results is the behavior of VAT101. In the homogeneous case, the combination of corner-adjusted PVM and transmissibility modifier produced the lowest RMSE among the uniform-type treatments (108.70 psi). In the heterogeneous reservoir, however, VAT101 overpredicts BHP throughout injection by 110 to 272 psi, reaches its peak deviation at year 9, and maintains a persistent offset of 72 psi at year 50. Its RMSE rises to 156.11 psi (NRMSE = 11.00\%), a substantial degradation relative to its homogeneous performance. This deterioration arises because the transmissibility modifier, derived from the harmonic mean of adjacent half-cell transmissibilities (Eq.~\ref{eq:tm_j}), assumes a uniform permeability field; when the boundary cells sample a heterogeneous permeability distribution, the modifier over-restricts flow across high-permeability faces and under-restricts it across low-permeability ones, producing a net throttling effect that inflates BHP. The corner-adjusted PVM-only case (VA101), which does not impose a transmissibility correction, avoids this: it underpredicts during early injection ($-98$ psi at year 2), overshoots modestly after shut-in ($+74$ psi near year 17), and converges to a terminal offset of only 10 psi, yielding an RMSE of 68.50 psi (NRMSE = 4.83\%). The reversal between VA101 and VAT101 relative to their homogeneous ranking is an important result of this study: in heterogeneous reservoirs, applying a uniform transmissibility correction can be worse than applying no transmissibility correction at all.

\begin{figure}[htbp]
    \centering
    \includegraphics[width=\textwidth]{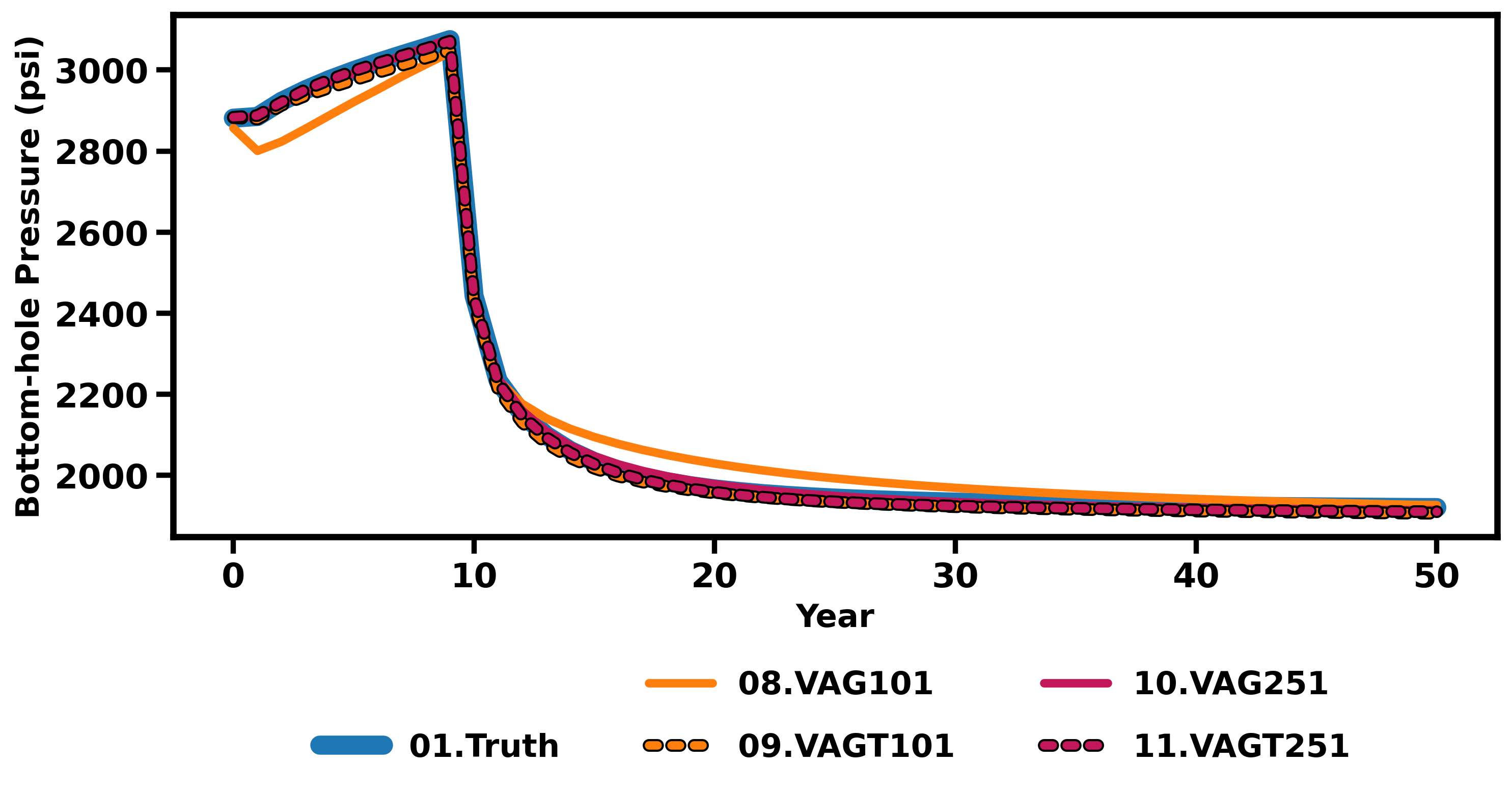}
    \caption{BHP comparison for the gradual power-law boundary treatments at two grid
    resolutions (heterogeneous model).}
    \label{fig:Figure9}
\end{figure}

The effect of enlarging the simulated area on the gradual modifier is shown in Figure~\ref{fig:Figure9}. On the smaller $101 \times 101$ area, VAG101 follows the familiar PVM-only pattern with peak deviations of $-100$ psi during injection and $+66$ psi post-injection, yielding an RMSE of 63.38 psi (NRMSE = 4.75\%). Adding the transmissibility gradient (VAGT101) reduces the RMSE to 51.85 psi (NRMSE = 3.65\%). Enlarging the simulated area to $251 \times 251$ cells produces a modest further improvement: VAG251 achieves an RMSE of 49.31 psi (NRMSE = 3.48\%) and VAGT251 reaches 47.40 psi (NRMSE = 3.34\%). Unlike the homogeneous case, where extending the area offered less to no accuracy gain, here the larger area yields a measurable, though still modest, reduction in RMSE. The improvement reflects the greater physical separation between the injection well and the boundary ring, which allows local pressure fluctuations generated by heterogeneous flow paths to dissipate before reaching the modified boundary cells.

\begin{figure}[htbp]
    \centering
    \includegraphics[width=\textwidth]{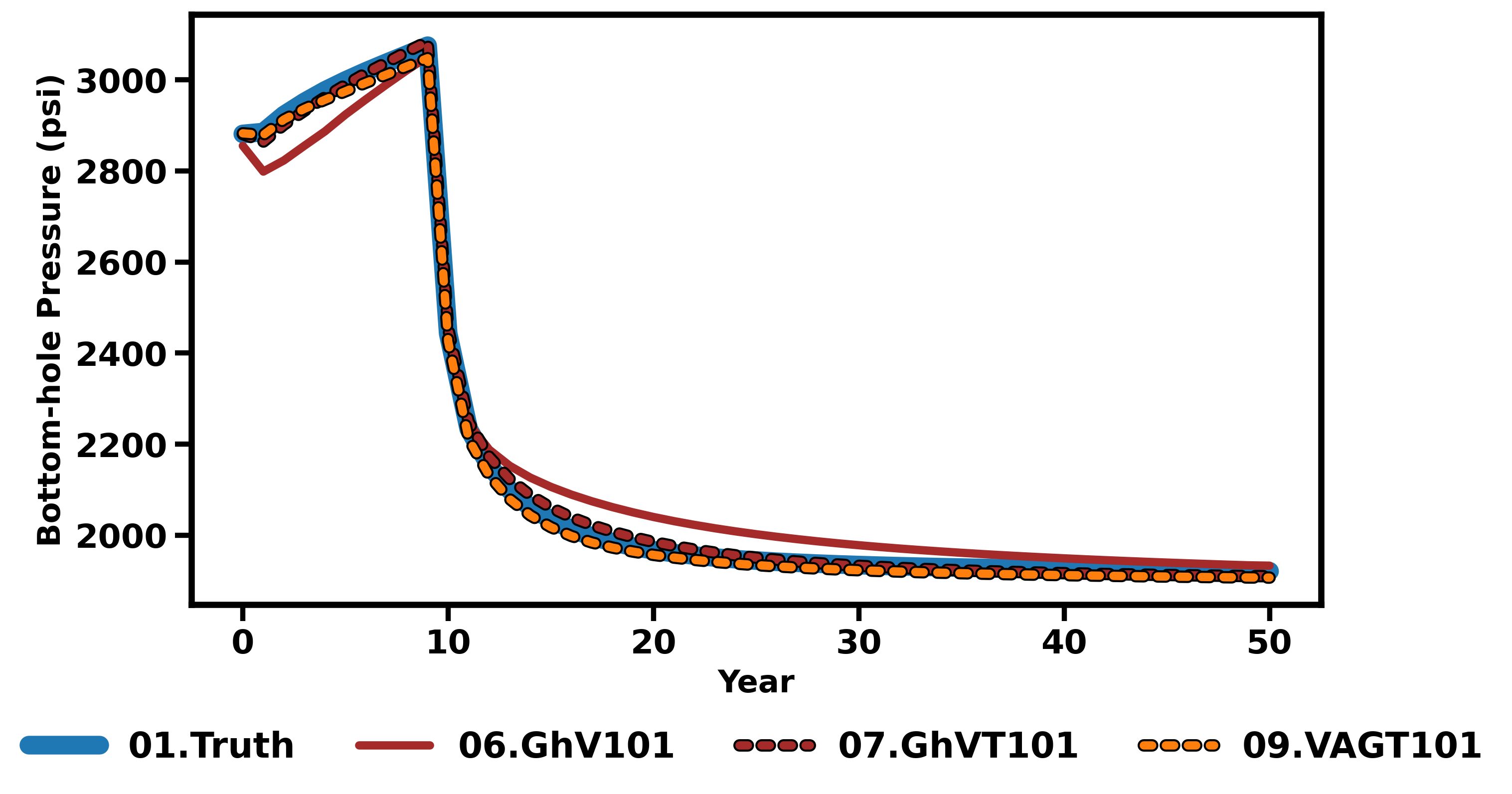}
    \caption{BHP comparison for the layered (GhV101, GhVT101) and gradual (VAGT101)
    boundary treatments on the $101 \times 101$ grid (heterogeneous model).}
    \label{fig:Figure10}
\end{figure}

The layered \parencite{GhomianEtAl2024} and gradual treatments on the $101 \times 101$ grid are compared in Figure~\ref{fig:Figure10}. GhV101 follows a trajectory nearly identical to VA101, underpredicting during injection ($-101$ psi at year 2) and overpredicting modestly post-injection ($+77$ psi near year 17), with an RMSE of 70.14 psi (NRMSE = 4.94\%). Adding the transmissibility correction (GhVT101) improves the match: the injection-phase deviation is compressed to $-22$ psi at year 2, the post-injection overshoot peaks at 27 psi near year 14, and the trajectory converges to within 11 psi of the truth by year 50. GhVT101 achieves an RMSE of 47.57 psi (NRMSE = 3.35\%).

The VAGT101 curve tracks the truth with comparable fidelity, underpredicting by up to 27 psi during late injection (year 9) and remaining within 14 psi of the truth after year 20. Its RMSE of 51.85 psi (NRMSE = 3.65\%) is close to that of GhVT101 (47.57 psi) and substantially lower than VAT101 (156.11 psi). The distributed-correction schemes, GhVT101 and VAGT101, both succeed in the heterogeneous reservoir because they spread the transmissibility reduction across multiple cells or smoothly along the boundary ring rather than imposing a single uniform throttle. This distribution avoids the abrupt property contrasts that amplify errors at high-permeability streaks intersecting the domain edge.

All heterogeneous RMSE results are summarized in Table~\ref{tab:rmse_hetero}. The most consequential finding relative to the homogeneous case is that the transmissibility modifier does not universally improve accuracy. VAT101 degrades from 3.48\% to 11.00\% NRMSE when heterogeneity is introduced, whereas the gradual (VAGT101) and layered (GhVT101) schemes remain effective because their distributed correction strategies are less sensitive to local permeability contrasts. The best-performing $101 \times 101$ configuration is GhVT101 (RMSE = 47.57 psi, NRMSE = 3.35\%), with VAGT101 (RMSE = 51.85 psi, NRMSE = 3.65\%) close behind, and the best overall is VAGT251 (RMSE = 47.40 psi, NRMSE = 3.34\%). Note that the \parencite{GhomianEtAl2024} approaches distribute the correction across four adjustment layers extending inward from the outer edge, allowing the scheme to more faithfully represent the heterogeneity present near the boundary, see Figure~\ref{fig:Figure4}c. Moreover, the narrow spread among the top distributed-correction schemes contrasts with the order-of-magnitude gap separating them from VAT101, we conclude that, for heterogeneous reservoirs, how the boundary correction is distributed matters more than the choice between layered and gradual implementations.

\begin{table}[htbp]
    \centering
    \caption{RMSE and NRMSE of BHP predictions for all boundary condition configurations
    on the heterogeneous model.}
    \label{tab:rmse_hetero}
    \begin{tabular}{lcc}
        \hline
        \textbf{Model} & \textbf{RMSE (psi)} & \textbf{NRMSE (\%)} \\
        \hline
        V101     & 250.11 & 17.63 \\
        VT101    & 304.17 & 21.44 \\
        VA101    &  68.50 &  4.83 \\
        VAT101   & 156.11 & 11.00 \\
        GhV101   &  70.14 &  4.94 \\
        GhVT101  &  47.57 &  3.35 \\
        VAG101   &  63.38 &  4.75 \\
        VAGT101  &  51.85 &  3.65 \\
        VAG251   &  49.31 &  3.48 \\
        VAGT251  &  47.40 &  3.34 \\
        \hline
    \end{tabular}
\end{table}

\subsection{CO\texorpdfstring{\textsubscript{2}}{2} Saturation Fidelity}

\subsubsection{Homogeneous}

CO$_2$ plume extent is assessed at the end of the 50-year simulation by comparing the gas-saturation footprint ($S_g > 0.001$) from each reduced-domain model with that of the truth model. Because the homogeneous reservoir has uniform porosity and permeability, the truth-model plume remains radially symmetric around the injection well. We use this symmetric plume as a clean baseline for isolating boundary distortions in the reduced-domain simulations.

\begin{figure}[htbp]
    \centering
    \includegraphics[width=0.75\textwidth]{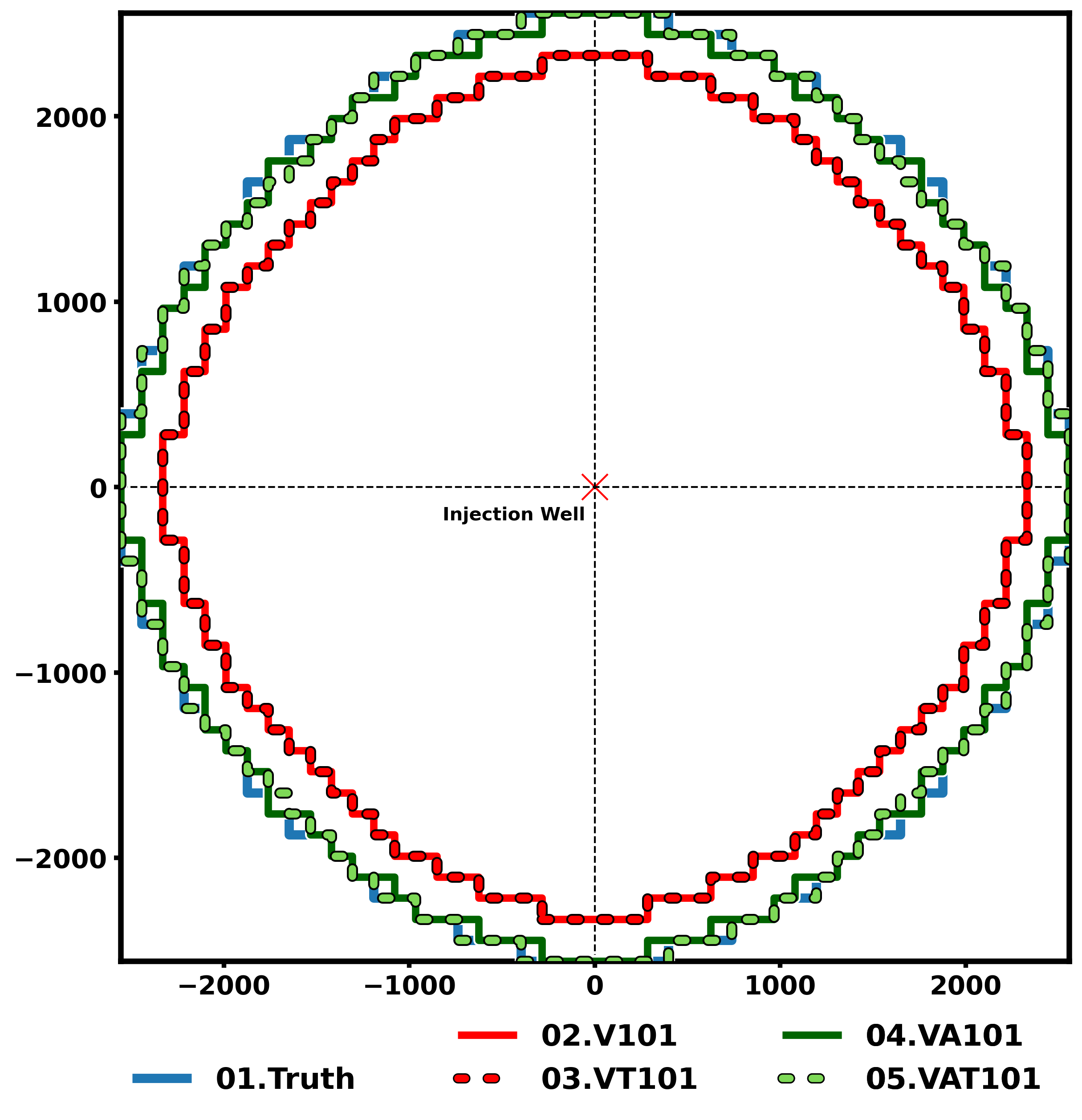}
    \caption{CO$_2$ plume outlines at year~50 for the uniform and corner-adjusted boundary
    treatments on the $101 \times 101$ grid (homogeneous model). Axes are in feet relative
    to the injection well.}
    \label{fig:Figure11}
\end{figure}

The plume outlines for the uniform and corner-adjusted treatments are shown in Figure~\ref{fig:Figure11}. V101 and VT101 both yield IoU values of 0.8021, indicating that roughly 20\% of the combined plume area is misrepresented relative to the truth (Table~\ref{tab:iou_homogeneous}). Despite their markedly different BHP trajectories during injection, the two cases produce identical plume footprints because the saturation front at year 50 has not reached the outermost cell ring; plume extent is governed by the interior grid, which is identical in both cases. The elevated BHP in VT101 does not translate into a larger plume because the transmissibility modifier restricts flow into the boundary cells, confining the CO$_2$ to a slightly smaller radial extent. The plume outlines of V101 and VT101 are visibly contracted relative to the truth, with the deficit most pronounced along all directions where the rectilinear grid discretization aligns the plume front with the boundary edges.

Corner adjustment recovers nearly all of the missing plume area. VA101 raises the IoU to 0.9780 and VAT101 to 0.9913, and their outlines closely track the truth in Figure~\ref{fig:Figure11}. The residual deviation is concentrated near the domain corners, where the circular plume front introduces a systematic geometric bias that no boundary modifier can fully eliminate on a Cartesian grid. The small advantage of VAT101 over VA101 (0.9913 versus 0.9780) is consistent with the transmissibility correction further refining the local pressure gradients near the boundary, although both schemes deliver IoU values within 2.3\% of unity and the practical distinction between them is minor for plume prediction in a homogeneous reservoir.

\begin{figure}[htbp]
    \centering
    \includegraphics[width=0.75\textwidth]{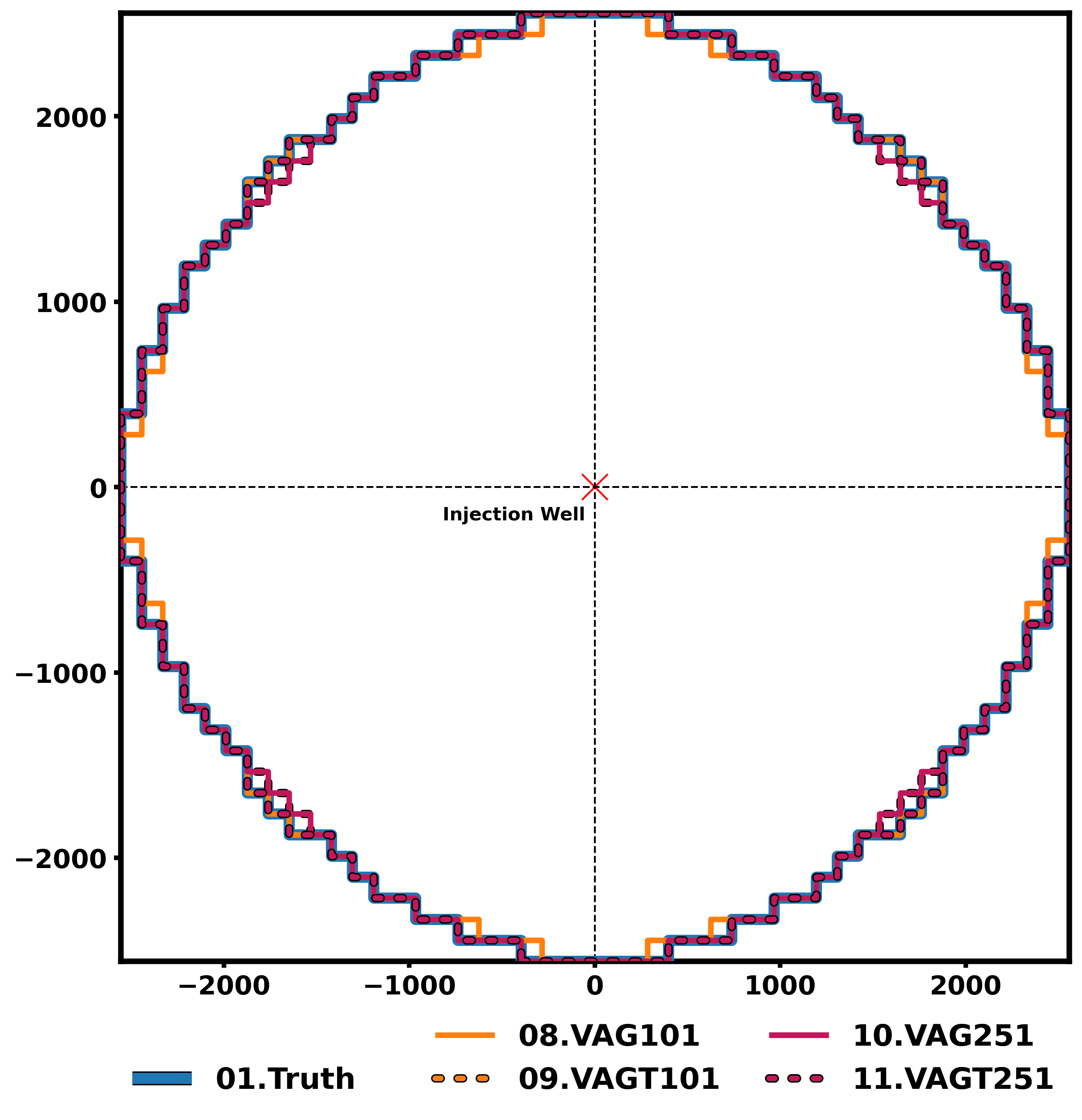}
    \caption{CO$_2$ plume outlines at year~50 for the gradual power-law boundary treatments
    at two grid resolutions (homogeneous model).}
    \label{fig:Figure12}
\end{figure}

The effect of enlarging the simulated area on the gradual modifier is presented in Figure~\ref{fig:Figure12}. VAG101 achieves an IoU of 0.9887, with visible shift from the truth near the domain corners where the plume front takes place. VAGT101 reaches 0.9913, marginally improving on VAG101 by smoothing the corner-region pressure gradients. Enlarging the area to $251 \times 251$ cells produces a modest further improvement: VAG251 attains 0.9913 and VAGT251 reaches 0.9993, with outlines that are virtually superimposed on the truth. The small residual deviations of VAG101 at the corners, visible as orange steps slightly inside the truth contour, vanish on the larger area of VAG251.

\begin{figure}[htbp]
    \centering
    \includegraphics[width=0.75\textwidth]{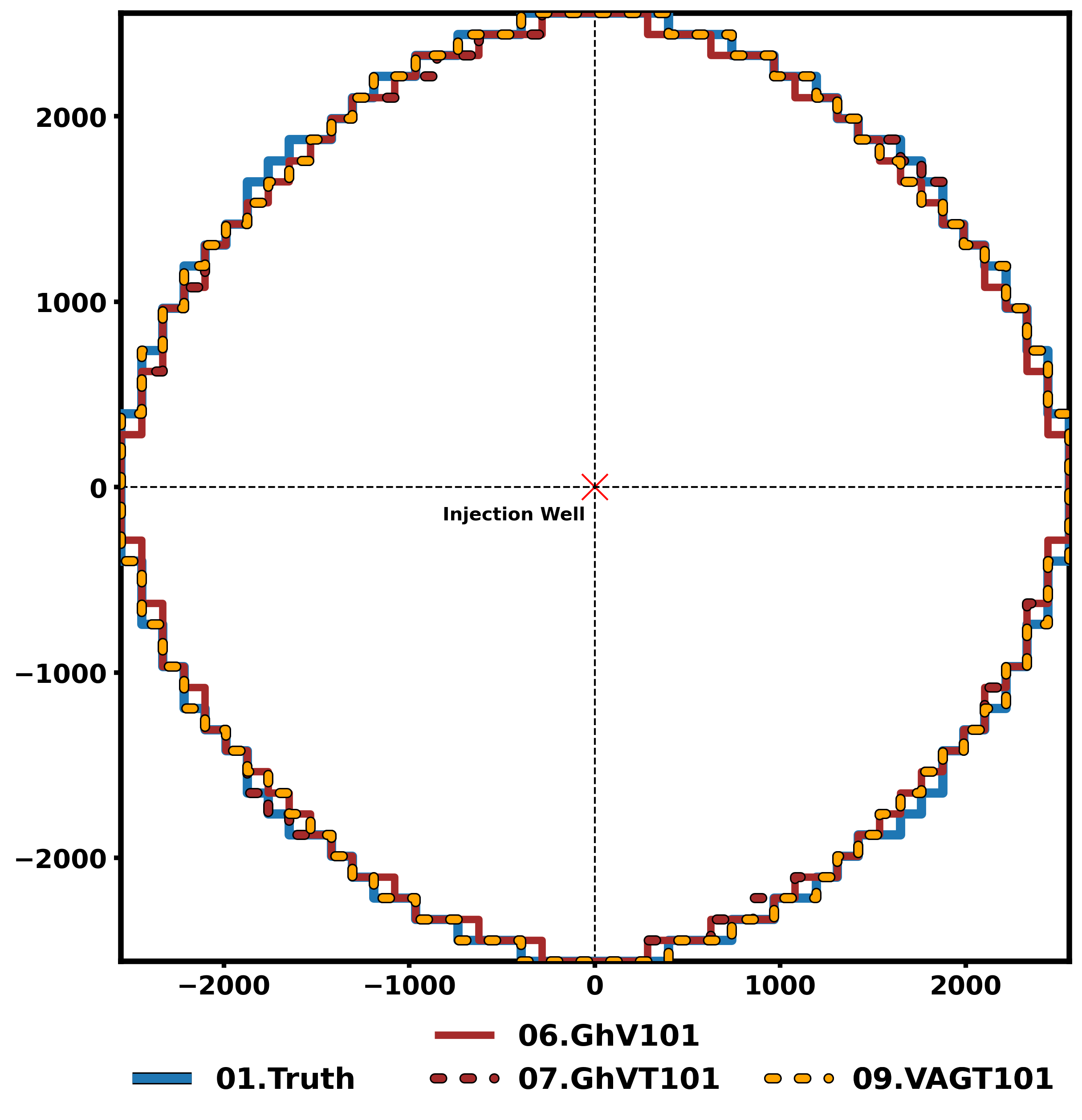}
    \caption{CO$_2$ plume outlines at year~50 for the layered (GhV101, GhVT101) and gradual
    (VAGT101) boundary treatments on the $101 \times 101$ grid (homogeneous model).}
    \label{fig:Figure13}
\end{figure}

The layered of \parencite{GhomianEtAl2024} and gradual treatments on the $101 \times 101$ grid are compared in Figure~\ref{fig:Figure13}. GhV101 yields an IoU of 0.9754, GhVT101 reaches 0.9873, and VAGT101 achieves 0.9913. Their outlines are largely superimposed on one another. The GhV101 plume exhibits minor local deviations from the truth near the upper-right and lower-left quadrants, where the four-layer PVM distribution creates subtle pressure gradients that slightly redirect the plume front; GhVT101 and VAGT101 smooth these irregularities. The three configurations are nearly indistinguishable in plume fidelity, indicating that once corner-volume conservation is achieved, the specific modifier distribution strategy exerts only a secondary influence on the saturation footprint in a homogeneous reservoir.

All configurations are summarized in Table~\ref{tab:iou_homogeneous}. The results divide cleanly into two groups. V101 and VT101, which lack corner-volume correction, both yield IoU values of 0.8021, while all remaining configurations cluster between 0.9754 and 0.9993 regardless of whether they employ layered, gradual, or uniform-with-corner boundary treatments. This clustering indicates that plume extent in the homogeneous reservoir is controlled primarily by whether the boundary layer conserves the total formation pore volume; the finer distinctions among modifier strategies that produced detectable differences in BHP accuracy have a negligible effect on the CO$_2$ saturation footprint.

\begin{table}[h]
\centering
\caption{Intersection over Union (IoU) of CO\textsubscript{2} plume at year 50 for all boundary condition configurations on the homogeneous model.}
\label{tab:iou_homogeneous}
\begin{tabular}{lc}
\hline
\textbf{Model} & \textbf{IoU} \\
\hline
Truth    & 1.0000 \\
V101     & 0.8021 \\
VT101    & 0.8021 \\
VA101    & 0.9780 \\
VAT101   & 0.9913 \\
GhV101   & 0.9754 \\
GhVT101  & 0.9873 \\
VAG101   & 0.9887 \\
VAGT101  & 0.9913 \\
VAG251   & 0.9913 \\
VAGT251  & 0.9993 \\
\hline
\end{tabular}
\end{table}

\subsubsection{Heterogeneous}

The heterogeneous reservoir introduces spatially correlated permeability contrasts that
break the radial symmetry of the plume, creating preferential flow paths along
high-permeability streaks and producing an irregular plume outline that is more sensitive to
boundary-induced pressure artifacts than its homogeneous counterpart.

\begin{figure}[htbp]
    \centering
    \includegraphics[width=0.75\textwidth]{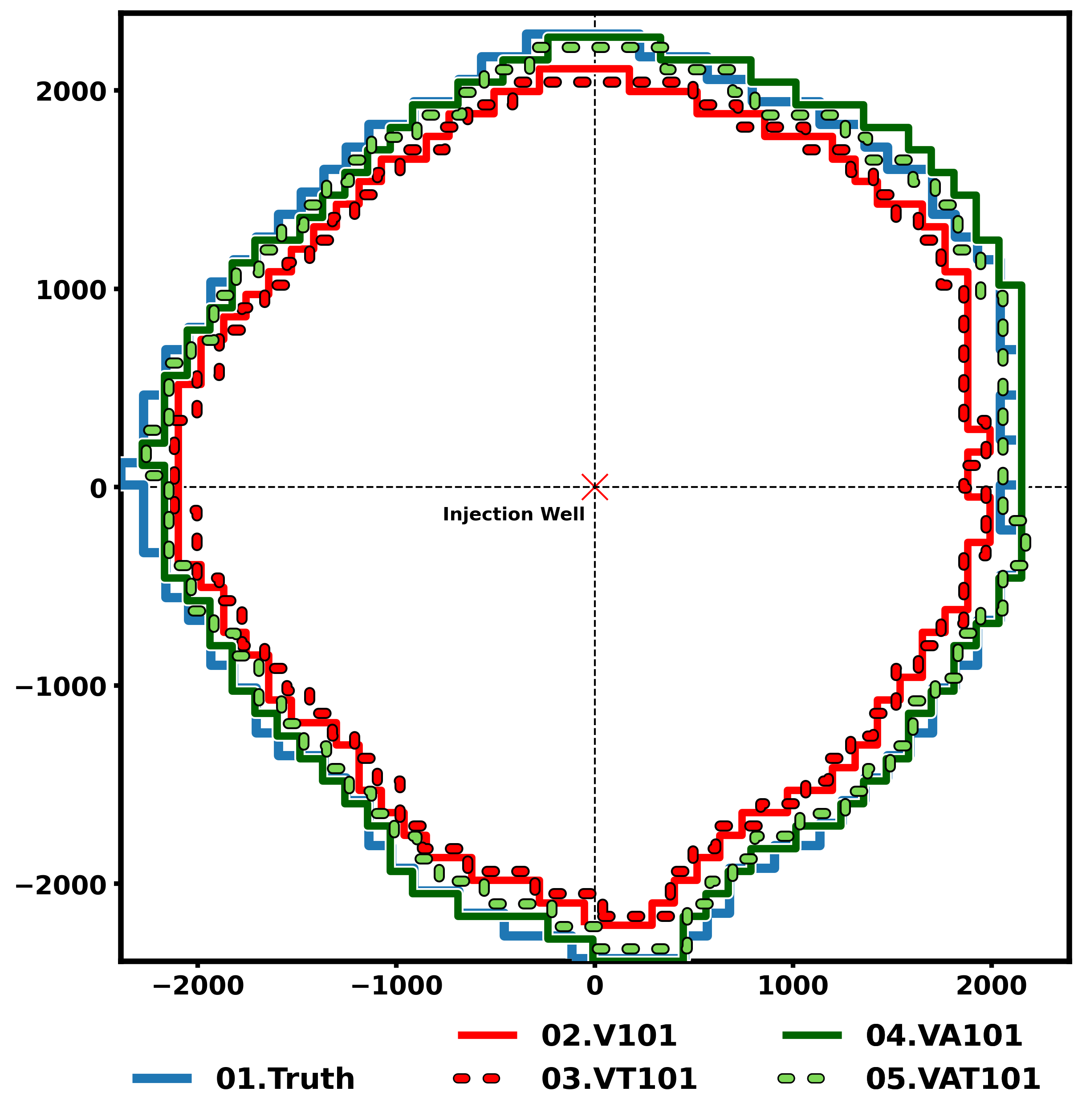}
    \caption{CO$_2$ plume outlines at year~50 for the uniform and corner-adjusted boundary
    treatments on the $101 \times 101$ grid (heterogeneous model). Axes are in feet relative
    to the injection well.}
    \label{fig:Figure14}
\end{figure}

The plume outlines for the uniform and corner-adjusted treatments are shown in Figure~\ref{fig:Figure14}. V101 and VT101 yield IoU values of 0.8439 and 0.8221, respectively (Table~\ref{tab:iou_heterogeneous}). Unlike the homogeneous case where these two configurations produced identical footprints, heterogeneity introduces a distinction: the transmissibility confinement in VT101 slightly alters the local pressure that steer the plume along permeability channels, yielding a marginally smaller footprint than V101 and a correspondingly lower IoU. Both outlines are visibly contracted relative to the truth across all quadrants, with the largest discrepancies along the left and lower boundaries where high-permeability pathways direct the plume front toward the domain edges.

Corner adjustment again recovers most of the missing plume area. VA101 raises the IoU to 0.9468 (Table~\ref{tab:iou_heterogeneous}). In Figure~\ref{fig:Figure14}, both VA101 and VAT101 track the truth more closely than V101 and VT101, although VA101 remains slightly inside the truth contour in several sectors. VAT101 yields an IoU of 0.9408, marginally below VA101 and a clear degradation relative to its homogeneous performance (0.9913). This parallels the BHP finding: the uniform transmissibility modifier over-restricts flow across heterogeneous boundary faces, and the resulting pressure inflation confines the CO$_2$ to a smaller and slightly displaced radial extent. The plume contraction is most visible in the right and lower-right quadrants of Figure~\ref{fig:Figure14}, where VAT101 falls noticeably inside the truth.

\begin{figure}[htbp]
    \centering
    \includegraphics[width=0.75\textwidth]{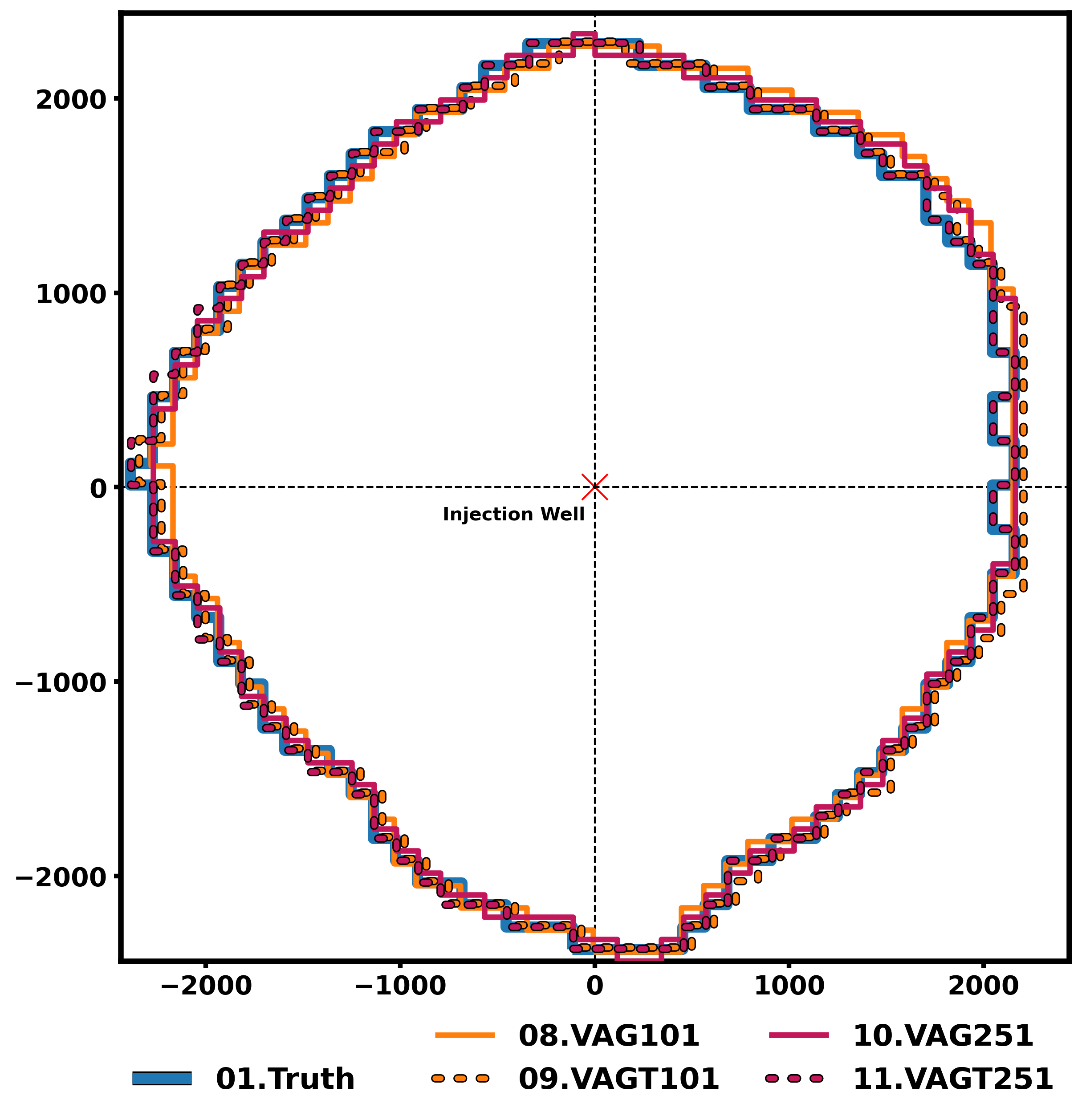}
    \caption{CO$_2$ plume outlines at year~50 for the gradual power-law boundary treatments
    at two grid resolutions (heterogeneous model).}
    \label{fig:Figure15}
\end{figure}

The effect of enlarging the simulated area on the gradual modifier is shown in Figure~\ref{fig:Figure15}. On the smaller $101 \times 101$ area, VAG101 achieves an IoU of 0.9473, with a visible contraction along the left boundary where a high-permeability streak directs the plume toward the domain edge. VAGT101 raises the IoU to 0.9723. Enlarging the area to $251 \times 251$ cells provides a meaningful further improvement: VAG251 reaches 0.9546 and VAGT251 attains 0.9862, with outlines nearly indistinguishable from the truth. The improvement on the larger area is more pronounced than in the homogeneous case, consistent with the BHP finding that additional buffer between the injection well and the boundary allows heterogeneity-driven pressure perturbations to dissipate before interacting with the modified boundary cells.

\begin{figure}[htbp]
    \centering
    \includegraphics[width=0.75\textwidth]{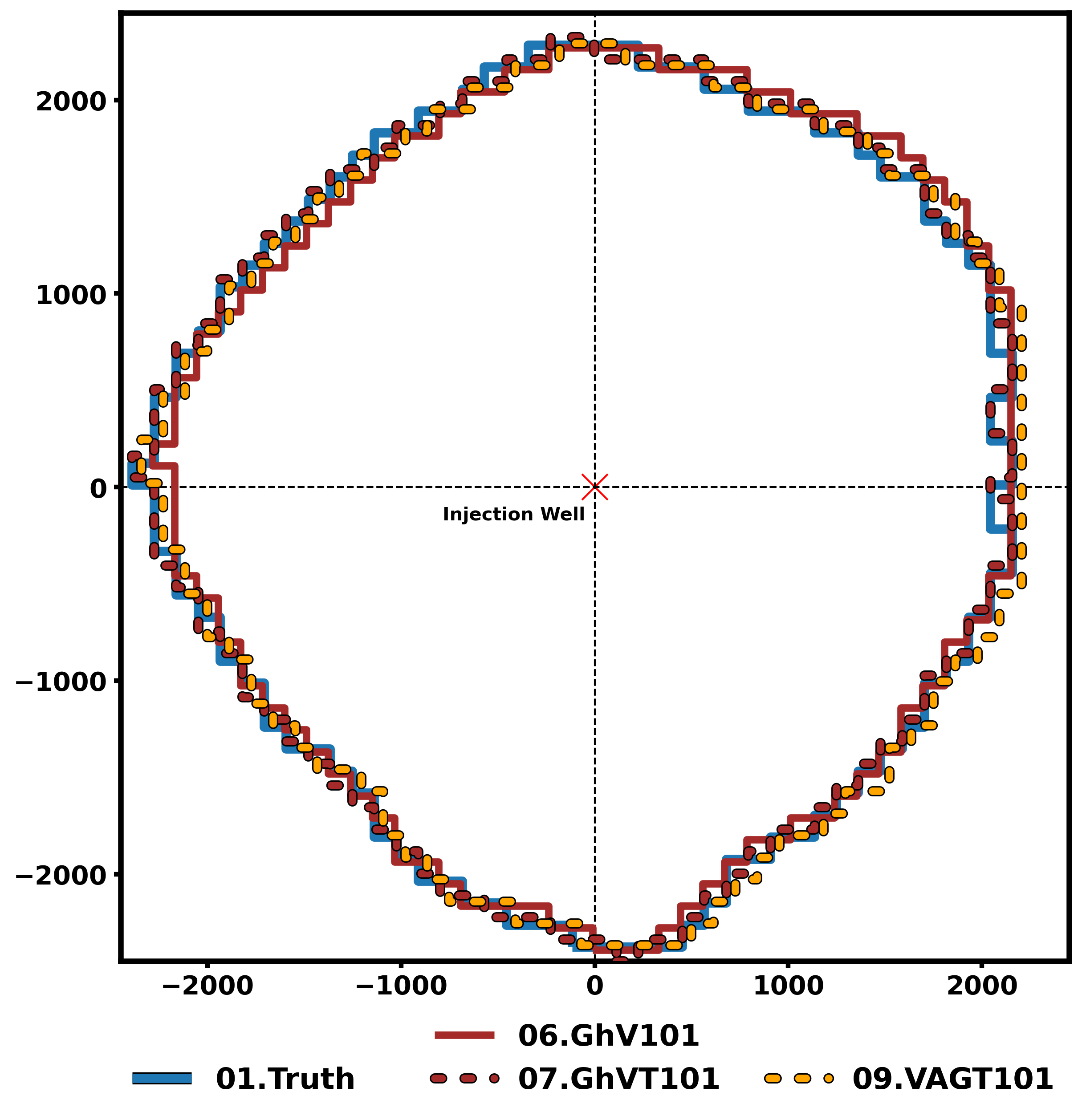}
    \caption{CO$_2$ plume outlines at year~50 for the layered (GhV101, GhVT101) and gradual
    (VAGT101) boundary treatments on the $101 \times 101$ grid (heterogeneous model).}
    \label{fig:Figure16}
\end{figure}

The layered and gradual treatments on the $101 \times 101$ grid are compared in Figure~\ref{fig:Figure16}. GhV101 yields an IoU of 0.9460, with its outline sitting slightly inside the truth in most sectors. Adding the transmissibility correction (GhVT101) raises the IoU to 0.9714; the GhVT101 contour extends beyond the truth in some locations and falls inside it in others, indicating that the four-layer transmissibility distribution generates local pressure that push the plume front outward in certain directions while pulling it inward in others, a sensitivity that was absent in the homogeneous case. VAGT101 achieves a similar IoU of 0.9723, very slightly above GhVT101. Its outline closely tracks the truth around the full perimeter, with minor shifts limited to isolated segments where permeability contrasts steer the plume front.

We compile all heterogeneous plume area deviations in Table~\ref{tab:iou_heterogeneous}. Heterogeneity amplifies the sensitivity of plume extent to the specific modifier strategy because boundary-induced pressure errors interact with preferential flow paths, redirecting the plume front in ways that a uniform reservoir does not permit. The clearest casualty is VAT101, which degrades from 0.9913 to 0.9408, consistent with its BHP deterioration and confirming that the uniform transmissibility modifier is unsuitable for heterogeneous reservoirs. The gradual modifier with transmissibility correction (VAGT101 and VAGT251) and the layered scheme with transmissibility correction (GhVT101) provide the most consistent plume fidelity across both reservoir architectures, with VAGT251 yielding the highest overall IoU at 0.9862.

\begin{table}[h]
\centering
\caption{Intersection over Union (IoU) of CO\textsubscript{2} plume at year 50 for all boundary condition configurations on the heterogeneous model.}
\label{tab:iou_heterogeneous}
\begin{tabular}{lc}
\hline
\textbf{Model} & \textbf{IoU} \\
\hline
Truth    & 1.0000 \\
V101     & 0.8439 \\
VT101    & 0.8221 \\
VA101    & 0.9468 \\
VAT101   & 0.9408 \\
GhV101   & 0.9460 \\
GhVT101  & 0.9714 \\
VAG101   & 0.9473 \\
VAGT101  & 0.9723 \\
VAG251   & 0.9546 \\
VAGT251  & 0.9862 \\
\hline
\end{tabular}
\end{table}
%% ============================================================
\section{Conclusions}
%% ============================================================

Ten boundary-condition configurations are evaluated against full-domain reference simulations for both homogeneous and heterogeneous reservoir models. BHP and CO$_2$ plume extent are used by us as co-equal validation criteria. The main findings are summarized as follows.

Corner-volume conservation is the most important correction. Uniform PVM schemes without corner scaling (V101 and VT101) lead to large BHP errors, with RMSE values of 362.14 and 382.35 psi in the homogeneous model and 250.11 and 304.17 psi in the heterogeneous model. We also observe CO$_2$ plume IoU values near 0.80 in the homogeneous case and between 0.82 and 0.84 in the heterogeneous case for these schemes, indicating that roughly 16 to 20\% of the combined plume area is misrepresented relative to the truth. By contrast, all corner-adjusted configurations reduce BHP RMSE by 60 to 72\% and raise plume IoU above 0.94 in both reservoir types, showing that proper treatment of corner cells is essential for preserving both pressure response and CO$_2$ plume extent.

The transmissibility modifier is not always beneficial. In the homogeneous reservoir, we find that VAT101 yields the lowest 101-cell BHP RMSE (108.70 psi, NRMSE = 3.48\%) and a plume IoU of 0.9913. In the heterogeneous case, however, its performance deteriorates: BHP NRMSE rises to 11.00\% and plume IoU drops to 0.9408. In heterogeneous reservoirs, that assumption can over-restrict flow across high-permeability boundary faces, leading to larger pressure and plume errors.

The gradual power-law modifier with transmissibility correction (VAGT) and the layered scheme with transmissibility correction (GhVT) deliver the most consistent accuracy across both reservoir types and both response variables. For VAGT101, we obtain BHP NRMSE of 3.56\% and plume IoU of 0.9913 in the homogeneous case, and BHP NRMSE of 3.65\% with plume IoU of 0.9723 in the heterogeneous case. Enlarging the simulated area to $251 \times 251$ cells, we find that VAGT251 reaches the highest overall plume fidelity, with IoU of 0.9993 in the homogeneous case and 0.9862 in the heterogeneous case.

Plume extent is less sensitive than BHP to the modifier strategy in homogeneous reservoirs, where we observe all corner-adjusted configurations clustering within a narrow IoU band of 0.9754 to 0.9993. In heterogeneous reservoirs this insensitivity breaks down, with IoU values spanning 0.8221 to 0.9862, confirming that boundary-induced pressure errors interact with preferential flow paths to redirect the plume front.

For practical GCS workflows, we recommend the corner-adjusted PVM (VA) as a robust low-complexity baseline for early screening, and VAGT as the default for detailed characterization and regulatory compliance modeling where both BHP and CO$_2$ plume fidelity are required.

\section*{Acknowledgement}
We gratefully acknowledge Computer Modelling Group Ltd. (Canada) for providing the CMG simulator licenses, and the Gulf Coast Carbon Center (GCCC) at the Bureau of Economic Geology (BEG), The University of Texas at Austin, for their support during manuscript preparation. This material is based  in part upon work supported  by the Department of Energy under DOE Award Number DE-FE0031558.

\section*{Disclaimer}
This report was prepared as an account of work sponsored by an agency of the United States Government. Neither the United States Government nor any agency thereof, nor any of their employees, makes any warranty, express or implied, or assumes any legal liability or responsibility for the accuracy, completeness, or usefulness of any information, apparatus, product, or process disclosed, or represents that its use would not infringe privately owned rights. Reference herein to any specific commercial product, process, or service by trade name, trademark, manufacturer, or otherwise does not necessarily constitute or imply its endorsement, recommendation, or favoring by the United States Government or any agency thereof. The views and opinions of authors expressed herein do not necessarily state or reflect those of the United States Government or any agency thereof.

\printbibliography

\end{document}